\documentclass{article}

\usepackage{microtype}
\usepackage{graphicx}
\usepackage{subcaption}
\usepackage{booktabs} % for professional tables
\usepackage{makecell}
\usepackage{hyperref}
\usepackage[accepted]{arxiv/icml2026}

\usepackage{amsmath}
\usepackage{amssymb}
\usepackage{mathtools}
\usepackage{amsthm}
\usepackage{tabularx}

\usepackage[table]{xcolor}

\usepackage[capitalize,noabbrev]{cleveref}

\theoremstyle{plain}

\theoremstyle{definition}

\theoremstyle{remark}

\usepackage{tcolorbox}

\usepackage{multirow} % 必选：用于合并行
\usepackage{array}    % 提供更强的列控制
\usepackage{colortbl} % 必选：为行/单元格着色
\usepackage{xcolor}   % 必选：定义高级颜色
\usepackage{pifont}   % 可选：用于添加小图标

\usepackage{tikz}

\usepackage{pgfplots}
\pgfplotsset{compat=1.18}

\usepgfplotslibrary{polar}
\pgfplotsset{compat=newest}
\usepackage{wrapfig}

\definecolor{color1}{RGB}{68, 114, 196}  % GSM8K
\definecolor{color2}{RGB}{237, 125, 49}  % MathVista
\definecolor{color3}{RGB}{165, 165, 165} % DocVQA
\definecolor{color4}{RGB}{255, 192, 0}   % CultureBench

\usetikzlibrary{positioning, arrows.meta} % 虽然当前代码没直接用，但在顶会绘图中常备

\definecolor{tableheadcolor}{gray}{0.92}
\definecolor{ourmethodcolor}{rgb}{0.9, 0.95, 1.0} % 淡淡的蓝色，突出自研方法
\definecolor{lightblue}{rgb}{0.933,0.968,0.988} % 附录
\definecolor{color_prompt}{RGB}{226,240,217} % 附录

\usepackage[textsize=tiny]{todonotes}

\usepgfplotslibrary{fillbetween}
\usetikzlibrary{intersections}

\begin{document}

\twocolumn[
    \icmltitle{Thinking with Comics: Enhancing Multimodal Reasoning through Structured Visual Storytelling}
  \icmlsetsymbol{equal}{*}

  \begin{icmlauthorlist}
  \icmlauthor{Andong Chen}{yyy}
\icmlauthor{Wenxin Zhu}{yyy}
\icmlauthor{Qiuyu Ding}{yyy}
\icmlauthor{Yuchen Song}{yyy}
\icmlauthor{Muyun Yang}{yyy}
\icmlauthor{Tiejun Zhao}{yyy}

\textbf{Website:} \url{https://thinking-with-comics.github.io/}

% \textbf{Repository:} \url{https://github.com/andongBlue/Think-with-Comics}.
    %\icmlauthor{}{sch}
    %\icmlauthor{}{sch}
  \end{icmlauthorlist}

  \icmlaffiliation{yyy}{Harbin Institute of Technology}
  % \icmlaffiliation{comp}{Company Name, Location, Country}
  % \icmlaffiliation{sch}{School of ZZZ, Institute of WWW, Location, Country}
  \icmlcorrespondingauthor{Andong Chen}{ands691119@gmail.com}

  \icmlcorrespondingauthor{Tiejun Zhao}{tjzhao@hit.edu.cn}

  % You may provide any keywords that you find helpful for describing your
  % paper; these are used to populate the "keywords" metadata in the PDF but
  % will not be shown in the document
  \icmlkeywords{Machine Learning, ICML}

  \vskip 0.3in
]

% this must go after the closing bracket ] following \twocolumn[ ...

% This command actually creates the footnote in the first column listing the
% affiliations and the copyright notice. The command takes one argument, which
% is text to display at the start of the footnote. The \icmlEqualContribution
% command is standard text for equal contribution. Remove it (just {}) if you
% do not need this facility.

% Use ONE of the following lines. DO NOT remove the command.
% If you have no special notice, KEEP empty braces:
\printAffiliationsAndNotice{}  % no special notice (required even if empty)
% Or, if applicable, use the standard equal contribution text:
% \printAffiliationsAndNotice{\icmlEqualContribution}

\begin{abstract}
Chain-of-Thought reasoning has driven large language models to extend from thinking with text to thinking with images and videos. However, different modalities still have clear limitations: static images struggle to represent temporal structure, while videos introduce substantial redundancy and computational cost.
In this work, we propose Thinking with Comics, a visual reasoning paradigm that uses comics as a high information-density medium positioned between images and videos. Comics preserve temporal structure, embedded text, and narrative coherence while requiring significantly lower reasoning cost. We systematically study two reasoning paths based on comics and evaluate them on a range of reasoning tasks and long-context understanding tasks. Experimental results show that Thinking with Comics outperforms Thinking with Images on multi-step temporal and causal reasoning tasks, while remaining substantially more efficient than Thinking with Video. Further analysis indicates that different comic narrative structures and styles consistently affect performance across tasks, suggesting that comics serve as an effective intermediate visual representation for improving multimodal reasoning.
\end{abstract}

\begin{figure*}[t]
  \centering
  \vskip 0.2in
  \includegraphics[width=\textwidth]{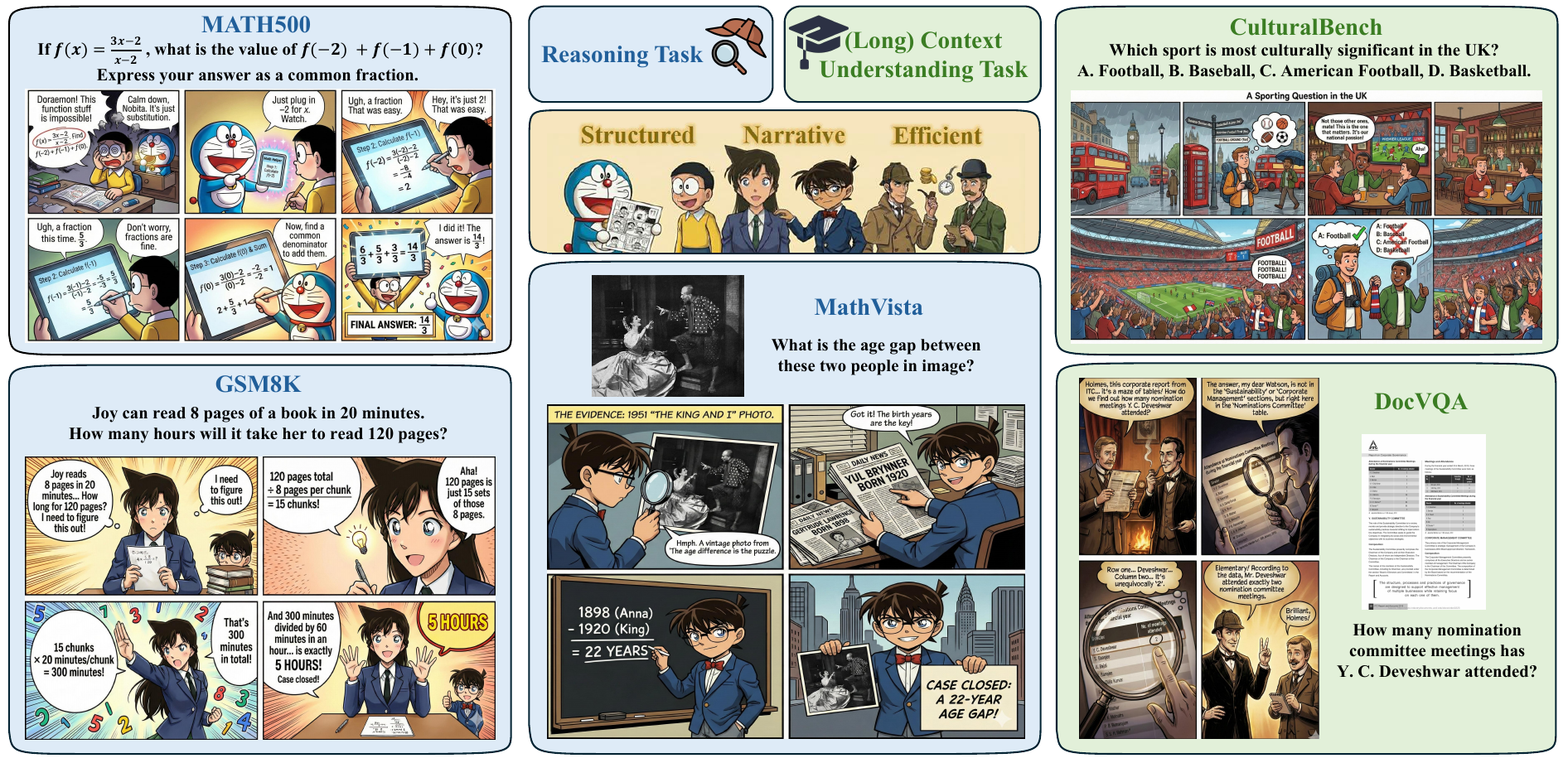}
  \caption{
    The selected reasoning tasks and (Long) Context Understanding tasks, along with the Thinking with Comics solution based on Gemini-3 Pro Image. The reasoning tasks primarily involve mathematical and logical reasoning, while the (Long) Context Understanding tasks require the model to comprehend cultural contexts, documents, and other extended information. The model provides the reasoning process and correct answers within the generated comic panels.
  }
  \label{intro_picture}
\end{figure*}

\section{Introduction}

% Chain-of-Thought (CoT) has significantly improved the reasoning abilities of large language models (LLMs)~\cite{wei2022chain}, making Thinking with Text a core reasoning paradigm. Models such as OpenAI o3, o4-mini, Gemini 3 Pro, and Janus-Pro can even Thinking with Images in their reasoning chains~\cite{hurst2024gpt,OpenAIOA}, meaning they not only view images but also generate them to support multimodal reasoning for VLMs. More recently, the idea of Thinking with Video~\cite{tong2025thinking} has emerged, where models generate short video sequences to enable human-like dynamic reasoning, thereby enhancing overall reasoning performance.
Large language models (LLMs) have significantly improved their reasoning ability on complex tasks by adopting explicit Chain-of-Thought (CoT)~\cite{wei2022chain, kojima2022large, besta2024graph, yao2023tree}, making step-by-step textual reasoning (Think with text) a common paradigm. With the development of multimodal large language models (MLLM), this idea of explicit reasoning has extended from pure text to the visual domain. Under the Thinking with Images (TWI) paradigm~\cite{hurst2024gpt,OpenAIOA, zhang2023multimodal, wang2025multimodal,chen2025make}, models not only use images as input signals but also generate intermediate visual representations during reasoning to supplement critical visual information~\cite{li2025imagine, hu2024visual}, thereby improving the reasoning performance of vision–language models (VLMs). Building on this, Thinking with Video further introduces temporal structure by generating short video sequences, enabling more complex forms of dynamic reasoning~\cite{tong2025thinking}.

Despite the extension of reasoning paradigms from text to images and videos, each modality still exhibits clear limitations. Static images struggle to represent temporal structure and dynamic processes, while the absence of explicit textual cues complicates cross-modal alignment. Videos provide temporal information but introduce substantial redundancy and significantly higher computational overhead, which limits their practical efficiency for reasoning.

% Despite the progress from text-based to video-based reasoning, each modality still has clear limitations. Static images cannot capture dynamic processes or temporal logic, and the absence of textual information makes it difficult to align understanding and generation across modalities. Videos, while containing temporal cues, introduce substantial redundancy and require significantly higher computational resources, creating bottlenecks in both reasoning accuracy and cost.

To address these limitations, we turn to a more natural reasoning medium from daily life-\textit{comics}-and introduce the \textbf{Thinking with Comics} (TwC) paradigm. Comics are a distinctive narrative form. Compared with static images, they retain most key properties of video, including temporal logic, embedded text, and dynamic reasoning~\cite{augereau2017overview}. Yet compared with video, each panel is more information-dense and requires far lower reasoning cost. Recent generative models such as Gemini-3 Pro Image~\cite{Gemini3Pro} can convert long text into coherent sequential panels while embedding text naturally within images. This allows comics to combine the high-density reasoning benefits of images with the dynamic logic of video. Thus, Thinking with Comics has strong potential to expand visual reasoning into a new research direction.

To comprehensively explore this field, we adopted two paths of Thinking with Comics, namely End-to-End Visualized Reasoning and Comic as Conditioning Context for VLM. Then we evaluate our method on mainstream general-purpose benchmarks across two task types, as shown in Figure~\ref{intro_picture}: (1) reasoning tasks and (2) (long) context understanding tasks. In the evaluation, we test the two paths and compare them with leading MLLMs as well as models that following the paradigms of Thinking with Text, Thinking with Images, and Thinking with Video. The results show that comics, as a form of structured visual storytelling, consistently yield systematic performance gains across different tasks.

Further analysis reveals that: (1) different tasks benefit from different role-playing narrative structures in comics—for example, detective-style narratives are better suited for logical reasoning tasks, while culture-centric narratives are more effective for cultural understanding; (2) Thinking with Comics exhibits scaling behavior similar to Chain-of-Thought, where more difficult tasks require a larger number of comic panels to support reasoning; (3) comic panels exhibit clear temporal and logical dependencies, and disrupting or permuting their order leads to noticeable performance degradation; (4) embedded textual elements in comics, such as dialogue and narration, work jointly with visual cues to reduce semantic ambiguity in purely visual reasoning; and (5) compared to Thinking with Video, Thinking with Comics achieves substantially lower inference cost while preserving essential temporal structure.

These findings indicate that visual expression still offers substantial room for exploration, and that comics provide a new reasoning medium positioned between static images and videos. We hope this work will inspire further exploration of Thinking with paradigms and help establish comics as an important component of a unified visual reasoning framework.

\section{Related Works}

\textbf{Reasoning Paradigm Transfer:} 
CoT enhances the interpretability of reasoning in LLMs by incorporating explicit intermediate reasoning steps, and significantly improves their reasoning capabilities~\cite{wei2022chain, kojima2022large, wang2022self, huang2023towards}. Inspired by this paradigm, some works have further introduced it into MLLMs, developing the Thinking with Images paradigm~\cite{hurst2024gpt, zhang2023multimodal, zheng2023ddcot, mitra2024compositional, gao2024cantor}, where MLLMs process original images or generate new ones and perform reasoning within an interleaved flow of textual and visual information. For both aforementioned paradigms, models typically employ large-scale reinforcement learning~\cite{shao2024deepseekmath, guo2025deepseek, liu2025visual} or some training-free inference-time scaling methods~\cite{kojima2022large, xu2025llava, dhuliawala2024chain} to enhance their CoT reasoning abilities. Recently, addressing issues in the Thinking with Images paradigm, such as the lack of temporal information in single image and the relative independence between textual and visual modalities, \citealp{tong2025thinking} proposed the Thinking with Video paradigm. This approach leverages video generation models like Sora 2 to integrate visual and textual reasoning within a unified temporal framework, where the video generation process itself constitutes the reasoning process.

\textbf{Vision Generation Model:}
The development of visual generation models has been profoundly influenced by diffusion models, which have become the mainstream methods for image and video generation~\cite{ho2020denoising}. A key milestone in this field is Stable Diffusion~\cite{rombach2022high}, a latent diffusion model used for efficiently generating high-resolution images. Building on these foundational architectures, the latest advances in image generation focus on enhancing text-to-image consistency, controllability, and fidelity. For example, DALL·E 3~\cite{betker2023improving} integrates advanced captioning and multimodal training to generate highly detailed and contextually accurate images from text prompts, addressing limitations in compositionality observed in earlier models. Similarly, Google's Nano Banana and its enhanced version Nano Banana Pro employ advanced image generation and editing techniques to achieve studio-level precise control and prompt accuracy, supporting natural language-described photo editing and high-quality image creation~\cite{Gemini3Pro}. Extending these principles to video generation, models such as OpenAI's Sora and its successor Sora 2 utilize spatiotemporal diffusion to generate coherent video sequences from text, incorporating world simulation capabilities to achieve realistic motion and long-range consistency~\cite{sora2}. Meanwhile, Google DeepMind's Veo 3 advances audiovisual generation by natively integrating sound effects and dialogue with high-fidelity video frames, utilizing 3D latent diffusion to enhance temporal coherence and multimodal expressiveness~\cite{veo3}. These models collectively represent a trajectory toward more versatile and integrated visual generative systems, paving the way for applications in creative industries and beyond.

\begin{figure*}[t]
  \centering
  \vskip 0.2in
  \includegraphics[width=\textwidth]{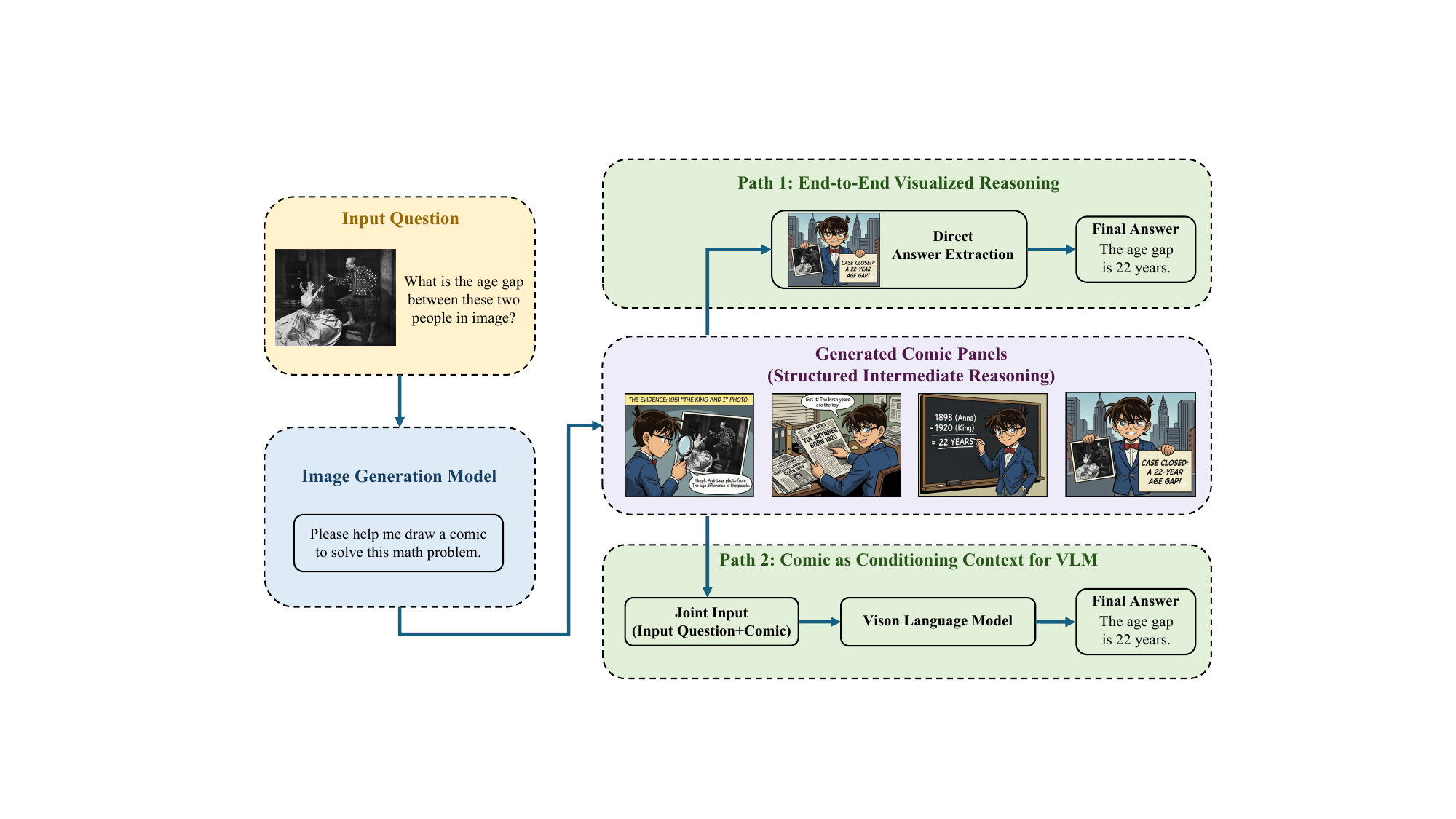}
  \caption{
    Overview of the two paths of Thinking with Comics paradigm. Path 1 directly utilizes an image generation model to create a comic, where the process of generating the comic constitutes the reasoning process for the problem, and the answer is obtained by extracting the final panel of the comic. Path 2 takes the generated comic along with the original problem as context and inputs them into a VLM, which then performs reasoning and outputs the answer.
  }
  \label{method_picture}
\end{figure*}

\section{Method}
\label{method}
In this section, we introduce Thinking with Comics, a novel structured visual storytelling reasoning paradigm that explicitly externalizes intermediate reasoning processes into a sequence of comic panels with temporal and causal structures. These panels serve either as the reasoning carrier itself or as conditioning context for downstream inference, enabling more interpretable and structurally grounded reasoning in multimodal models.
 
From the implementation perspective, Thinking with Comics can be instantiated through two paths. As shown in Figure~\ref{method_picture}, the first path treats comic generation as the reasoning process itself, where a generative model performs end-to-end visualized reasoning from the input question to the final answer. The second path instead regards the generated comic as an explicit intermediate reasoning representation, which is then combined with the original question and processed by a MLLM for joint reasoning. In the following, we describe these two paths in detail.

 % and analyze their differences in terms of reasoning mechanism, information flow, and applicability.

\subsection{Path $\mathrm{I}$: End-to-End Visualized Reasoning}
% The first path is conceptually related to the Thinking with Video paradigm, in which models such as Sora 2 generate a reasoning video whose temporal evolution implicitly represents the reasoning trajectory, and the final answer is obtained by processing the last video frame. In a similar approach, we employ an image generation model like Gemini-3 Pro Image to generate a comic that visually contains the reasoning process of the input question, and we extract the answer from the last panel of the comic.

The first path uses an image generation model to produce a comic based on the input question, visually depicting the reasoning process, and extracts the final answer from the last panel of the comic.

Formally, let \( q \in \mathcal{Q} \) denote the input question, and let \( \theta \) be the parameters of the image generation model. The model generates a sequence of comic panels \( \mathcal{C} = {c_1, c_2, \dots, c_T} \), where each panel \( c_t \) depicts an intermediate reasoning step. The generation process is expressed as:
\begin{equation}
    \mathcal{C} = G_{\theta}(q).
\end{equation}
During generation, the model progressively unfolds the reasoning process, with each panel corresponding to a reasoning state. In this path, reasoning and generation are tightly coupled. We assume that the model implicitly learns a latent state transition process:
\begin{equation} 
    h_t = f(h_{t-1}, q), \quad c_t = g(h_t),
\end{equation}
where \( h_t \) denotes the latent reasoning state at step \( t \), and \( g(\cdot) \) maps the latent state to a visual comic panel.
The final answer \( \hat{a} \) is obtained by extracting information from the last panel:
\begin{equation}
    \hat{a} = R(c_T),
\end{equation}
where \( R(\cdot) \) denotes an answer extraction process that identifies relevant textual or symbolic information from the final panel.

This path provides an end-to-end reasoning framework with relatively low computational cost, while offering interpretable intermediate representations. The sequential and causally coherent nature of comic panels enables the reasoning trajectory to be directly visualized. However, since all reasoning is performed implicitly within the generation model, the overall reasoning capability is constrained by the model itself.
% , making it difficult to enhance logical rigor or generalization ability through external reasoning components.

\subsection{Path $\mathrm{II}$: Comic as Conditioning Context for VLM}
The second path treats comics as an explicit intermediate reasoning medium and incorporates a MLLMs for downstream inference. This design is related to image-assisted reasoning approaches in the Thinking with Images paradigm, while providing a more structured and temporally consistent representation through multi-panel comics.

In this path, a comic is first generated though image generation model:
\begin{equation}
    \mathcal{C} = G_{\theta}(q),
\end{equation}
and the original question \( q \) together with the comic \( \mathcal{C} \) are then provided as input to a MLLMs:
\begin{equation}
    \hat{a} = F_{\phi}(q, \mathcal{C}),
\end{equation}
where \( F_{\phi} \) denotes a MLLMs parameterized by \( \phi \).
To formalize the influence of comics in the reasoning process, we treat the comic as an explicit intermediate variable \( z \): 
% , yielding:
\begin{equation}
    z = \mathcal{C}, \quad \hat{a} = \arg\max_a p(a \mid q, z).
\end{equation}
\vskip -3pt
Compared to textual intermediate variables used in traditional CoT reasoning, the comic representation \( z \) jointly encodes spatial structure, object relationships, and temporal evolution. This richer representation provides the MLLMs with a structured and multimodal reasoning context.

\begin{table*}[!ht]
\centering
\small
\setlength{\tabcolsep}{4pt}
\renewcommand{\arraystretch}{1.2}
\caption{Main results on reasoning and context understanding benchmarks. \textbf{M-Vista} and \textbf{Cultu.} denote MathVista and CulturalBench, respectively. G-t-R is Generate-then-Reason. For CulturalBench, \textbf{E} and \textbf{H} represent the Easy and Hard subsets. The symbol “---” indicates that the model does not support the specific task. * denotes results from \citealp{tong2025thinking}; $\star$ indicates evaluation on 50 sampled instances, following \citealp{tong2025thinking}.}

\scalebox{0.98}{
\begin{tabular}{ll c ccc c cc}
\toprule
\multirow{2}{*}{\textbf{Category}} & \multirow{2}{*}{\textbf{Model / Method}} & \multirow{2}{*}{\textbf{Notes}} & \multicolumn{3}{c}{\textbf{Reasoning Benchmarks (Acc \%)}} & & \multicolumn{2}{c}{\textbf{Context Understanding (Acc \%)}} \\
\cmidrule(lr){4-6} \cmidrule(lr){8-9}
& & & \textbf{MATH-500} & \textbf{GSM8K} & \textbf{M-Vista} & & \textbf{DocVQA} & \textbf{Cultu. (E / H)} \\
\midrule
\multirow{3}{*}{MLLM} 
& GPT-5.2 & direct & 99.0 & 100.0 & 67.5 & & 72.8 & 88.3 / 84.4 \\
& Gemini-3-Pro & direct & \textbf{100.0} & 99.0 & 71.5 & & 94.5 & \textbf{90.4} / \textbf{90.0} \\
& Claude-Sonnet 4.5 & direct & 99.0 & 100.0 & 72.5 & & 92.6 & 87.2 / 76.5 \\
\midrule
\multirow{2}{*}{Reasoning LLM} 
& DeepSeek-R1 & CoT & 90.4 & 96.1 & --- & & --- & 87.2 / 85.1 \\
& Qwen3-235B-A22B & CoT & 92.4 & 94.3 & --- & & --- & 83.1 / 82.5 \\
\midrule
\multirow{2}{*}{Think with Image} 
& TWI-1-Generated Photo & G-t-R & 70.2 & 69.4 & 63.6 & & 67.5 & 69.7 / 71.4 \\
% & TWI-Method-2 & G-t-R & 66.6 & 66.6 & 66.6 & & 66.6 & 66.6 / 66.6 \\
& DREAMLLM & G-t-R & 12.6 & 18.4 & 35.9 & & 65.5 & 52.3 / 42.8 \\
\midrule
Think with Video & Sora 2 & V-o-T & 67.0* & 75.7* & $67.6^\star$ & & $50.5^\star$ & $60.0^\star$ / $70.0^\star$ \\
\midrule
\multirow{2}{*}{\textbf{Think with Comic}} 
& \textbf{TwC (Ours) - Only Image} & direct & 90.0 & \textbf{100.0} & 75.0 & & 92.8 & 70.0 / 80.5 \\
& \textbf{TwC (Ours) - Img \& Txt} & G-t-R & 92.3 & 95.4 & \textbf{85.8} & & \textbf{99.4} & 88.3 / 82.2 \\
\bottomrule
\end{tabular}}
\label{tab:main_results_1}
\end{table*}

\section{Experiments}

\subsection{Evaluation Datasets}

% \textbf{Evaluation Datasets.}
We evaluate the proposed Thinking with Comics on a diverse set of benchmarks covering both explicit reasoning and multimodal understanding capabilities. The evaluation datasets are grouped into two task categories: reasoning tasks and (long) context understanding tasks.

The reasoning tasks include MATH500~\cite{lightman2023let}, GSM8K~\cite{cobbe2021training}, and MathVista~\cite{lu2023mathvista}, which primarily require multi-step logical or mathematical inference. MATH500 and GSM8K focus on symbolic and numerical reasoning in purely textual settings, while MathVista extends these challenges to visually grounded mathematical problems that demand joint visual perception and logical reasoning.

% The (long) context understanding tasks consist of eBDtheque~\cite{guerin2013ebdtheque}, DocVQA~\cite{mathew2021docvqa}, and CulturalBench~\cite{chiu2024culturalbench}, and aim to evaluate a model’s ability to understand and integrate contextual information beyond explicit step-by-step reasoning. eBDtheque is used for comic translation task and test the model's narrative comprehension ability and visual-text alignment ability across multiple panels; DocVQA focuses on document-level understanding over complex layouts and long contexts; and CulturalBench is a purely textual benchmark that probes culturally grounded and context-dependent semantic understanding. Together, these benchmarks emphasize sensitivity to narrative, document, and cultural context, rather than explicit logical inference.
(Long) context understanding tasks include DocVQA~\cite{mathew2021docvqa}, eBDtheque~\cite{guerin2013ebdtheque}, and CulturalBench~\cite{chiu2024culturalbench}. DocVQA primarily evaluates a ability to aggregate and understand document-level inputs; eBDtheque, designed for comic translation, focuses on document-level multilingual understanding and visual–text alignment across multiple panels; and CulturalBench is a text-only benchmark with two subsets (Easy / Hard) for evaluating contextualized cultural understanding. Overall, these benchmarks emphasize sensitivity to long documents, narrative structure, and cultural context, rather than explicit logical reasoning.

\subsection{Models and Experimental Setup.}
% \textbf{Models and Setup.}
In the experiments, we evaluate implementation paths of the Thinking with Comics paradigm introduced in \cref{method}.

For path $\mathrm{I}$ (End-to-End Visualized Reasoning), we directly employ  Gemini-3 Pro Image~\cite{Gemini3Pro}~\footnote{https://deepmind.google/models/gemini-image/pro/} to generate comics conditioned on the input question. The generated comic serves as the complete reasoning trajectory, and the final answer is extracted from the last panel.

For path $\mathrm{II}$ (Comic as Conditioning Context for MLLM), we first use Gemini-3 Pro Image to generate a comic, which is then provided together with the original question as input to a MLLM for joint reasoning. For convenience, we choose Gemini-3 Pro~\cite{Gemini3Pro} for further reasoning.

Unless otherwise specified, all models are evaluated in a zero-shot setting. Prompt templates are designed to ensure fair comparison across different reasoning paradigms while avoiding task-specific tuning.

\textbf{Baselines.}
We compare against four groups of strong baselines, including several frontier models: (i) \textit{text-only MLLMs}, including GPT-5.2~\cite{singh2025openai}, Gemini 3 Pro~\cite{Gemini3Pro}, and Claude Sonnet 4.5~\cite{ClaudeSonnet45}, which perform reasoning without explicit intermediate reasoning process~\footnote{The versions of the three MLLMs are respectively: gpt-5.2-2025-12-11, gemini-3-pro-preview, and claude-sonnet-4-5-20250929.}; (ii) \textit{Reasoning-oriented LLMs}, including DeepSeek-R1~\cite{guo2025deepseek}, Qwen3-235B-A22B~\cite{yang2025qwen3}, which are specifically designed to enhance multi-step reasoning through structured or implicit Chain-of-Thought mechanisms~\footnote{The version of two reasoning LLMs are respectively: deepseek-r1-0528 and qwen3-235b-a22b-thinking-2507}; (iii) \textit{models following the “Thinking with Images” paradigm}, including prompt-based approaches such as G-IMG~\cite{cheng2025visual}(the prompt provided in the Appendix~\ref{G-IMG}), as well as training-based methods like DREAMLLM~\cite{dong2023dreamllm}. These models assist reasoning by incorporating image generation or image-conditioned inputs during inference, with DREAMLLM relying on end-to-end training with a 7B-scale model.; and (iv) \textit{models following the Thinking with Video paradigm}, represented by Sora 2, where temporal video generation implicitly encodes the reasoning process.

\textbf{Metrics.}
For most benchmarks, we adopt accuracy as the evaluation metric, including MATH500, GSM8K, MathVista, DocVQA and CulturalBench, which directly measures the correctness of the final predicted answers. The details of answer extraction for the two TwC pathways are provided in Appendix~\ref{exact_final_answer_appendix}.
% For eBDthequ, we formulate it as a comic translation task and evaluate model performance using BLEU, which quantifies the n-gram overlap between the generated translations and the reference texts, providing a standard and widely used measure for translation quality.

\subsection{Main Results}

Table~\ref{tab:main_results_1} summarizes our systematic evaluation of TwC across reasoning benchmarks (MATH-500, GSM8K, MathVista) and context understanding benchmarks (DocVQA and CulturalBench). The results show that TwC performs strongly on multimodal reasoning tasks, achieving 85.8\% accuracy on MathVista and significantly outperforming Thinking with Video. On pure text-based mathematical reasoning benchmarks, TwC remains competitive with strong proprietary models. For context understanding tasks, TwC reaches 99.4\% accuracy on DocVQA and achieves leading performance on CulturalBench, particularly on the hard subset. Overall, these results demonstrate that introducing comic-style reasoning processes not only enhances both textual and visual reasoning, but also generalizes effectively to diverse context understanding tasks, validating the soundness and generalization capability of the TwC paradigm.

\section{Analysis Experiment}

\subsection{Role-playing Narrative Alignment}

We investigate how specific Role-playing narrative frameworks—such as documentary-style, detective-style, and slice-of-life comic pictures—serve as “Role-playing Narratives” to induce specific reasoning paths in path $\mathrm{I}$ of TwC. We compare three comic-mediated styles (documentary, detective, and slice-of-life) on the MathVista and GSM8K benchmarks, and observe performance variance when the model handles complex spatial and logical deduction tasks. The prompts and examples for each style are provided in Appendix~\ref{appendix_style_compare_prompt} and ~\ref{appendix_style_compare_example}.

\begin{table}[!ht]
\centering
\small
\caption{Narrative Style Ablation on MathVista and GSM8K. \textbf{Detective} style acts as the most effective visual prompt for tasks.}
\label{tab:style_alignment}
\begin{tabularx}{\columnwidth}{X c c c} % X用于自动填充，c用于数据居中
\toprule
\textbf{Style (Visual Prompt)} & \textbf{M-Vista} & \textbf{GSM8K} & \textbf{Avg. $\Delta$} \\
\midrule
Documentary (Base) & 60.0 & 68.0 & --- \\
Slice-of-Life      & 80.0 & 86.3 & +19.1 \\
\rowcolor[gray]{0.9} 
\textbf{Detective Style} & \textbf{85.0}* & \textbf{100.0}* & \textbf{+28.5} \\
\bottomrule
\end{tabularx}
\end{table}
\vskip -1pt
% \begin{table}[!ht]
% \centering
% \small % 缩小字体以适应单栏
% \caption{Narrative Style Ablation on MathVista and GSM8K. \textbf{Detective} style acts as the most effective visual prompt for reasoning.}
% \scalebox{0.98}{
% \begin{tabularx}{\columnwidth}{lccc} % 自动适配单栏宽度
% \toprule
% \textbf{Style (Visual Prompt)} & \textbf{M-Vista} & \textbf{GSM8K} & \textbf{Avg. $\Delta$} \\
% \midrule
% Documentary (Base) & 60.0 & 68.0 & --- \\
% Slice-of-Life    & 80.0 & 86.3 & +19.1 \\
% \rowcolor[gray]{0.9} % 浅灰色背景突出重点
% \textbf{Detective Style} & \textbf{85.0}* & \textbf{100.0}* & \textbf{+9.3} \\
% \bottomrule
%     \end{tabularx}}
% \label{tab:style_alignment}
% \end{table}
% \vskip -1pt
% vskip
% \vspace{-0.5cm}

% Experimental results, as shown in Table \ref{tab:style_alignment}, reveal that the detective-style significantly outperforms the standard documentary-style comic in logical reasoning tasks, with a 28.5-point accuracy gain, corresponding to a 44.5\% relative improvement over the documentary baseline. 

Experimental results, as shown in Table \ref{tab:style_alignment}, reveal that the detective‑style significantly outperforms the standard documentary‑style comic in logical reasoning tasks. Averaged across the two benchmarks, accuracy increases from (60.0 + 68.0)/2 = 64.0 to (85.0 + 100.0)/2 = 92.5, yielding a 28.5‑point absolute gain. This corresponds to a relative improvement of 28.5 / 64.0 = 44.5\% over the documentary baseline.
This suggests that role-playing narrative style is not merely a visual decoration but a potent Visual System Prompt. The results confirm that specific role-playing narrative structures established via comic panels can effectively activate the potential of MLLM for causal reasoning, leading to a more focused inference path. Appendix~\ref{appendix_photo_comics} analyzes the advantages of comic narratives over realistic-style images, comparing full comics with interleaved realistic image sequences in reasoning coherence and information organization.

\subsection{Scaling the Panels}
This experiment explores the scaling law of reasoning capability by varying the number of generated panels ($N \in \{1, 2, 4, 6, 8\}$) in the path $\mathrm{I}$ of TwC. Note that $N=1$ represents a degeneration into the traditional Think with Image (TWI) mode. We record the accuracy and token consumption when solving complex MATH500 problems to quantify the information compression efficiency of comics.

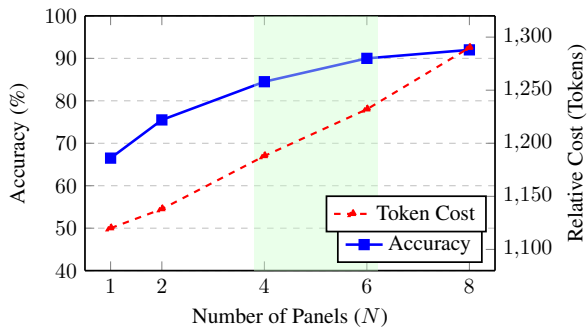
\begin{figure}[!ht]
\centering
\scalebox{0.82}{
\begin{tikzpicture}
\begin{axis}[
    width=0.48\textwidth,
    height=0.25\textheight,
    xlabel={Number of Panels ($N$)},
    ylabel={Accuracy (\%)},
    xmin=0.5, xmax=8.5,
    ymin=40, ymax=100,
    xtick={1,2,4,6,8},
    ytick={40,50,60,70,80,90,100},
    legend pos=south east,
    ymajorgrids=true,
    grid style=dashed,
    axis y line*=left, % 左轴：Accuracy
    thick,
]
% --- Accuracy Curve ---
\addplot[
    color=blue,
    mark=square*,
    line width=1.2pt,
]
coordinates {
    (1, 66.5) (2, 75.5) (4, 84.5) (6, 90.0) (8, 92.0)
};
\addlegendentry{Accuracy}

% 标注 Pareto 最优区域
\draw[fill=green!20, opacity=0.4, draw=none] (axis cs:3.8,40) rectangle (axis cs:6.2,100);
\node[text=green!60!black, font=\small\bfseries] at (axis cs:5, 88) {};

\end{axis}

\begin{axis}[
    width=0.48\textwidth,
    height=0.25\textheight,
    hide x axis,
    ylabel={Relative Cost (Tokens)},
    xmin=0.5, xmax=8.5,
    % --- 调整：允许轻微波动但保持整体上升趋势 ---
    ymin=1080, ymax=1320,                 % 略微放宽范围以显式呈现非线性小幅波动
    ytick={1100, 1150, 1200, 1250, 1300},  % 保持可读刻度
    axis y line*=right,
    legend style={at={(0.98,0.15)}, anchor=south east},
    thick,
]
% --- Cost Curve ---
\addplot[
    color=red,
    mark=triangle*,
    dashed,
    line width=1pt,
]
coordinates {
    (1, 1120)
    (2, 1138)   
    (4, 1188)   
    (6, 1232)  
    (8, 1290)  
};
% --- 修改部分结束 ---
\addlegendentry{Token Cost}

\end{axis}
\end{tikzpicture}}
\caption{The performance-cost curve across different panel counts $N$. Accuracy enters a plateau at $N \in [4, 6]$. On the MATH500 dataset, token cost ranges between 1100 and 1300.}
\label{fig:scaling_law}
\end{figure}
\vskip -10pt

As illustrated in the performance-cost curve in Figure~\ref{fig:scaling_law}, reasoning accuracy enters a visible plateau at 4--6 panels, while marginal gains from increasing panels diminish rapidly. The experimental results demonstrate that comics capture dynamic logic with minimal redundancy through high-level abstraction of continuous temporality. We conclude that 4--6 panels represent the Pareto optimal state between information density and computational overhead.

\subsection{Panel Distribution Across Task Difficulties}
This experiment counts the number of generated panels across different difficulty levels to reveal the adaptive mechanism of TwC. We analyzed thousands of samples from GSM8K (basic logic), MathVista (visual reasoning), DocVQA (long-document understanding), and CulturalBench-hard (cultural understanding). The model decides the number of panels based on the complexity of the problem. This tests if the model can allocate visual resources dynamically according to task difficulty.

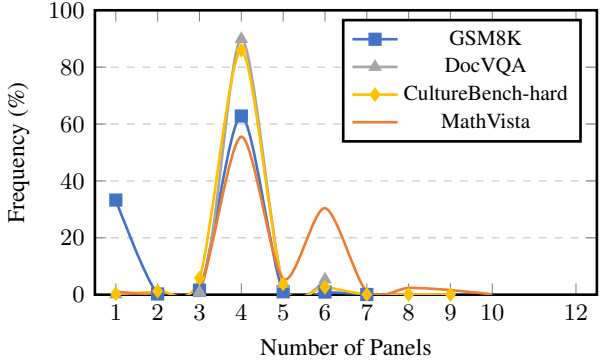
\begin{figure}[hbt!]
\centering
\begin{tikzpicture}
\begin{axis}[
    % --- 核心调整区域 ---
    width=\linewidth,      % 将宽度设置为当前栏目的宽度
    height=0.65\linewidth, % 高度按比例设置（0.6到0.7之间通常视觉效果较好）
    line width=1pt,        % 稍微减细线条以适应小图 (原为1.2pt)
    label style={font=\small},       % 缩小坐标轴标签字体
    tick label style={font=\footnotesize}, % 缩小刻度数字字体
    legend style={font=\small, nodes={scale=0.9, transform shape}}, % 缩小图例并紧凑化
    % --------------------
    xlabel={Number of Panels},
    ylabel={Frequency (\%)},
    xmin=0.5, xmax=12.5,
    ymin=0, ymax=100,
    xtick={1,2,3,4,5,6,7,8,9,10,12},
    ytick={0,20,40,60,80,100},
    legend pos=north east,
    ymajorgrids=true,
    grid style=dashed,
    smooth, % 使曲线平滑
]

% GSM8K - 明显的左偏态，Panel=1 占比极高
\addplot[color=color1, mark=square*] coordinates {
    % (1,62) (2,22) (3,10) (4,4) (5,2)
    (1,33.28) (2,0.25) (3,1.61) (4,62.82) (5,1.02) (6,0.93) (7,0.08)
};
\addlegendentry{GSM8K}

% DocVQA - 中等难度，分布在 2-4
\addplot[color=color3, mark=triangle*] coordinates {
    % (1,5) (2,25) (3,42) (4,20) (5,8)
    (1,0.11) (3,0.91) (4,89.89) (5,3.75) (6,5.34)
    
};
\addlegendentry{DocVQA}

% CultureBench-hard - 较高难度，中心在 4-5
\addplot[color=color4, mark=diamond*] coordinates {
    % (2,5) (3,15) (4,35) (5,28) (6,12) (7,5)
    (1,0.33) (2,1.23) (3,5.83) (4,85.95) (5,3.78) (6,2.63) (7,0.08) (8,0.08) (9,0.08)
};
\addlegendentry{CultureBench-hard}

% MathVista - 高阶推理，中心右移至 5-6，方差更大
\addplot[color=color2, mark=circle*] coordinates {
    % (2,2) (3,8) (4,15) (5,26) (6,24) (7,14) (8,8) (9,3)
    (1,1.08) (2,0.11) (3,1.84) (4,55.52) (5,5.63) (6,30.41) (7,1.08) (8,2.38) (9,1.62) (10,0.11)
};
\addlegendentry{MathVista}

\end{axis}
\end{tikzpicture}
\caption{Frequency distribution of generated panels across tasks with varying difficulty levels. The shift to the right indicates the model's adaptive allocation of reasoning steps for complex tasks.}
\label{fig:panel_distribution}
\end{figure}

Results are visualized in Figure~\ref{fig:panel_distribution}. GSM8K exhibits a bimodal distribution: while a substantial portion of easier samples (33.28\%) are efficiently solved with a single panel, the majority (62.82\%) still utilize 4 panels. In contrast, MathVista demonstrates a higher hard reasoning task; although also peaking at 4 panels, its distribution significantly extends towards higher panel counts, with a notable 30.41\% of samples requiring 6 panels. These shifts confirm that TwC allocates minimal resources (1 panel) for simple queries while dynamically extending reasoning for more complex tasks like MathVista.

% $\mathcal{S} = (p_1, p_2, \dots, p_N)$
\subsection{The Role of Temporal Sequence in Reasoning}
To examine whether the model captures temporal relationships across panels rather than relying on single-image features, we conduct a controlled logic test on path $\mathrm{II}$ of TwC by systematically perturbing the temporal structure of comic panel sequences. We design two controlled groups: \textit{Complete Shuffle} and \textit{Random Intermediate Deletion} observing model performance in MATH500 step-by-step solutions and comic translation tasks. Formally, given an ordered panel sequence \( \mathcal{C} = {c_1, c_2, \dots, c_T} \), we define the shuffle intensity $\sigma \in [0,1]$ as the proportion of panels whose temporal positions are permuted:
\begin{equation}
\sigma = \frac{1}{T} \sum_{i=1}^{T} \mathbb{I}[\pi(i) \neq i],
\end{equation}
where $\pi$ denotes a random permutation of panel indices. Here, $\sigma = 0$ corresponds to the original generated comic, while $\sigma = 1$ denotes \textit{Complete Shuffle}.
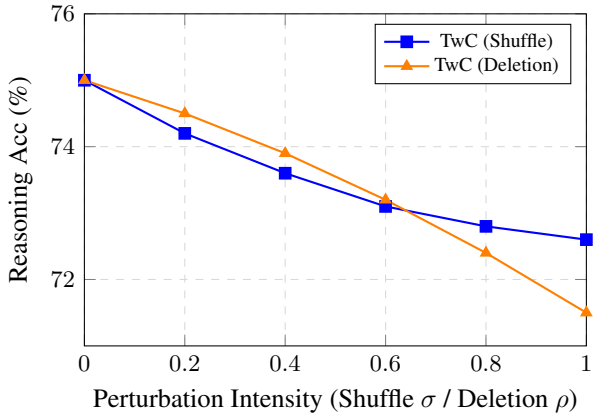
\begin{figure}[!ht]
    \centering
    \begin{tikzpicture}
        \begin{axis}[
            width=\linewidth,
            height=6cm,
            % --- 核心修改 1: 将 X 轴标签通用化 ---
            xlabel={Perturbation Intensity (Shuffle $\sigma$ / Deletion $\rho$)},
            ylabel={Reasoning Acc (\%)},
            xmin=0, xmax=1,
            ymin=71, ymax=76,
            xtick={0, 0.2, 0.4, 0.6, 0.8, 1.0},
            grid=both,
            grid style={dashed, gray!30},
            legend pos=north east,
            legend style={nodes={scale=0.75, transform shape}}, % 图例稍微缩小以容纳更多项
            tick label style={font=\small}
        ]
        
      % --- 曲线 1: TwC (Shuffle) ---
        \addplot[color=blue, mark=square*, thick] coordinates {
            (0,   75.0)
            (0.2, 74.2)
            (0.4, 73.6)
            (0.6, 73.1)
            (0.8, 72.8)
            (1.0, 72.6)
        };
        \addlegendentry{TwC (Shuffle)}
        
        % --- 曲线 2: TwC (Deletion) ---
        \addplot[color=orange, mark=triangle*, thick] coordinates {
            (0,   75.0)
            (0.2, 74.5)
            (0.4, 73.9)
            (0.6, 73.2)
            (0.8, 72.4)
            (1.0, 71.5)
        };
        \addlegendentry{TwC (Deletion)}
        
        % --- 曲线 3: Baseline (对照组) ---
        % 如果 Baseline 也做了两种测试，你可以画两条灰线，或者保留一条平均线
        % \addplot[color=gray, dashed, mark=o, thick] coordinates {
        %     (0, 55.0) (0.2, 54.2) (0.4, 52.8) (0.6, 51.5) (0.8, 50.2) (1.0, 49.8)
        % };
        % \addlegendentry{Multi-Image Baseline}
        
        % Annotation (可选，为了不遮挡新线，可以调整位置或去掉)
        % \draw[stealth-stealth, red, thick] (axis cs: 0, 88.2) -- (axis cs: 1, 42.1) 
        %    node[midway, sloped, above, font=\small] {$\Delta > 40\%$};
            
        \end{axis}
    \end{tikzpicture}
    % \caption{\textbf{Effect of Temporal Structure on Reasoning.} Model accuracy under two controlled manipulations of comic panel sequences: \textit{Complete Shuffle} (blue), which disrupts temporal order, and \textit{Intermediate Deletion} (orange), which removes intermediate panels while preserving relative order. The x-axis indicates the intensity of the manipulation. Accuracy consistently decreases as temporal structure is increasingly disturbed, with intermediate deletion causing a larger degradation than shuffling, highlighting the importance of temporal sequence information in comic-based reasoning.}
    \caption{Effect of temporal perturbations on comic-based reasoning. Accuracy under \textit{Complete Shuffle} (blue) and \textit{Intermediate Deletion} (orange) decreases as perturbation intensity increases, with deletion causing a larger drop than shuffling.}

    \label{fig:robustness_curves}
\end{figure}
\vskip -5pt

For \textit{Random Intermediate Deletion}, we randomly remove a subset of panels while preserving the relative order of the remaining ones. The deletion ratio $\rho \in [0,1]$ is defined as:
\begin{equation}
\rho = \frac{|\mathcal{D}|}{T},
\end{equation}
where $\mathcal{D} \subset \mathcal{C}$ denotes the set of deleted panels.

Experimental data in Figure~\ref{fig:robustness_curves} show that under Shuffle and Deletion conditions, the model's accuracy exhibits a decline from 75.0\% to 71.5\%.  These results verify that the model depends on the temporal logic across panels, rather than treating them as isolated images. Notably, missing temporal sequence information harms the reasoning process more than disordered inputs.

% This resilience provides strong evidence that the TwC paradigm encodes robust \textit{Gutter Reasoning}—the logic between panels. The findings indicate that the model effectively utilizes semantic white spaces (gutters) to anchor the logical loop, capturing temporal fluidity in a way that remains stable even under structural perturbation.

% Experimental data in Figure~\ref{fig:robustness_curves} show that under the Shuffled condition, the model’s accuracy in causal judgment tasks exhibits a consistent but moderate decline as perturbation intensity increases. This behavior provides evidence that the core advantage of the TwC paradigm lies in its ability to exploit \textit{Gutter Reasoning}—the relational logic between panels. The results indicate that the model actively leverages semantic white spaces (gutters) to maintain causal continuity, enabling stable reasoning under structural disturbances in a manner that traditional static multi-image settings fail to support.

\subsection{Ablation on Textual Anchoring}
This experiment quantifies the contribution of embedded textual elements—such as speech bubbles, narration, and onomatopoeia—in eliminating visual ambiguity and enhancing semantic comprehension. In Path~$\mathrm{II}$, we perform an ablation study on CulturalBench and MathVista, comparing \textit{pure visual panels} with comics containing complete \textit{bubbles and symbols}. We focus on the speed at which textual signals complement visual cognition in highly coupled scenarios. The prompts for each style are provided in Appendix \ref{appendix_text_anchor_prompt}.

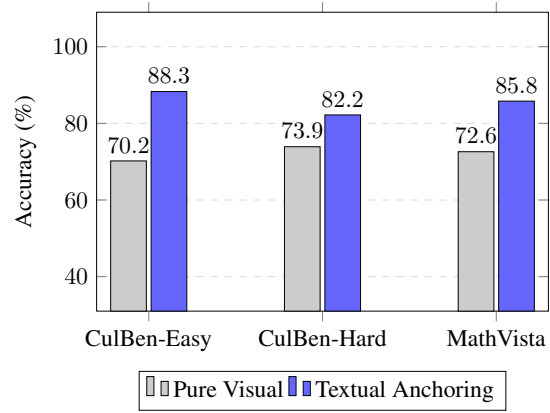
\begin{figure}[!ht]
    \centering
    \begin{tikzpicture}[scale=0.9]
        \begin{axis}[
            ybar,
            enlargelimits=0.15,
            legend style={at={(0.5,-0.2)}, anchor=north, legend columns=-1},
            ylabel={Accuracy (\%)},
            symbolic x coords={CulBen-Easy, CulBen-Hard, MathVista},
            xtick=data,
            nodes near coords,
            nodes near coords align={vertical},
            % ylim={40, 100},
            ymin=40, 
            ymax=100,
            width=\columnwidth,
            height=6cm,
            bar width=15pt,
            ymajorgrids=true,
            grid style={dashed, gray!30}
        ]
            % Pure Visual Data
            \addplot[fill=gray!40] coordinates {(CulBen-Easy, 70.2) (CulBen-Hard, 73.9) (MathVista, 72.6) };
            % With Text Anchoring Data
            \addplot[fill=blue!60] coordinates {(CulBen-Easy, 88.3) (CulBen-Hard, 82.2) (MathVista, 85.8)};
            
            \legend{Pure Visual, Textual Anchoring}
        \end{axis}
        % % Annotation for 15% Gain
        % \draw[<->, red, thick] (5.8, 2.7) -- (5.8, 3.5) node[midway, left] {\scriptsize +21.6};
        
    \end{tikzpicture}
    \caption{Ablation results on textual anchoring. Embedded text (bubbles, narration) provides precise semantic cues.}
    \label{fig:ablation_text}
\end{figure}
\vskip -5pt

% The comparative analysis in Figure~\ref{fig:ablation_text} shows that in cultural and long context understanding tasks, embedded text improves reasoning accuracy. In particular, the performance gap on DocVQA reaches 21.6 points. These results confirm that speech bubbles serve a \textit{Semantic Anchoring} role in comic contexts, eliminating image polysemy through precise linguistic instructions. This textual and visual modality integration significantly reduces the complexity of searching for correct solutions within the cross-modal space.
% The comparative results in Figure~\ref{fig:ablation_text} show that embedded text significantly improves reasoning accuracy in cultural understanding and reasoning-related tasks. In particular, performance on DocVQA increases by 21.6 points. The comparative analysis in Figure~\ref{fig:ablation_text} shows that in cultural and long context understanding tasks, embedded text improves reasoning accuracy. In particular, the performance gap on DocVQA reaches 21.6 points.

As shown in Figure~\ref{fig:ablation_text}, comics with embedded text consistently outperform pure visual panels across all evaluated tasks. Textual anchoring yields an accuracy gain of 18.1 points on CulturalBench-Easy, 8.3 points on CulturalBench-Hard, and 13.2 points on MathVista. These results confirm that speech bubbles serve a \textit{Semantic Anchoring} role in comic contexts, eliminating image polysemy through precise linguistic instructions. This textual and visual modality integration significantly reduces the complexity of searching for correct solutions within the cross-modal space.

\subsection{Cross-Model Generalization}

This experiment evaluates the cross-model generalization of path $\mathrm{II}$ in the TwC paradigm across diverse MLLMs architectures. We use the same TwC generated comic as a unified input and conduct large-scale evaluations on Claude 3.7 Sonnet, Qwen-VL-72B, GPT-5.2, Gemini 3 Pro, and GPT-4o~\footnote{The versions of these models are respectively: claude-3-7-sonnet-20250219, qwen2.5-vl-72b-instruct, gpt-5.2-2025-12-11, gemini-3-pro-preview, and gpt-4o-2024-05-13.}. The evaluation covers four capability categories and five benchmarks: logical reasoning (MATH-500, GSM8K), visual reasoning (MathVista), cultural understanding (CulturalBench), and long document understanding (DocVQA). By comparing model performance under an identical comic path, we assess TwC’s potential as a model-agnostic visual reasoning plug-in in terms of transferability and stability.

% \begin{table}[!ht]
% \centering
% \caption{Cross-Model Generalization Performance (ACC).}
% \label{tab:cross_model}
% \setlength{\tabcolsep}{3.5pt} % 减小列间距
% \footnotesize % 使用较小字号适配单栏
% \begin{tabular}{l|ccccc}
% \toprule
% \textbf{Dataset} & \textbf{\begin{tabular}[c]{@{}c@{}}Claude\\ 3.7\end{tabular}} & \textbf{\begin{tabular}[c]{@{}c@{}}Qwen\\ 72B\end{tabular}} & \textbf{\begin{tabular}[c]{@{}c@{}}GPT\\ 4o\end{tabular}} & \textbf{\begin{tabular}[c]{@{}c@{}}GPT\\ 5.2\end{tabular}} & \textbf{\begin{tabular}[c]{@{}c@{}}Gemini\\ 3 Pro\end{tabular}} \\ \midrule
% CB-Easy   & 88.25\% & 88.25\% & \textbf{88.41\%} & 88.10\% & 87.94\% \\
% CB-Hard   & 81.80\% & \textbf{82.21\%} & \textbf{82.21\%} & 81.97\% & \textbf{82.21\%} \\
% gsm8k     & 86.10\% & 75.21\% & 80.90\% & 85.01\% & \textbf{95.39\%} \\
% MATH-500  & 85.21\% & 84.51\% & 85.92\% & 85.92\% & \textbf{92.25\%} \\
% MathVista & 85.45\% & 85.88\% & \textbf{86.85\%} & \textbf{86.85\%} & 85.78\% \\
% DocVQA    & \textbf{99.66\%} & 99.44\% & 99.44\% & 99.55\% & 99.44\% \\ \bottomrule
% \end{tabular}
% \end{table}

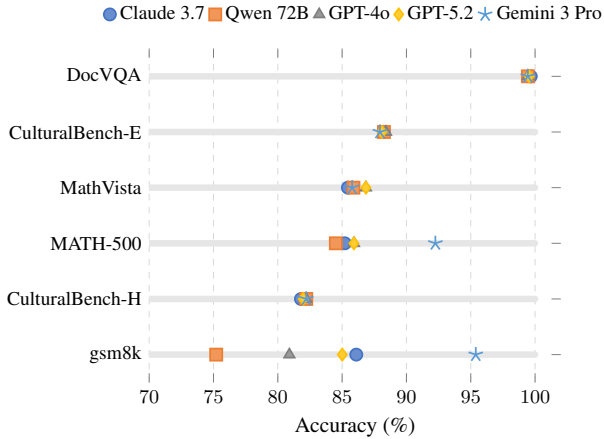
\begin{figure}[!ht]
    \centering 
    % --- 定义易于区分的学术配色 ---
    \definecolor{c1}{RGB}{68, 114, 196}  % Claude - 蓝
    \definecolor{c2}{RGB}{237, 125, 49}  % Qwen - 橙
    \definecolor{c3}{RGB}{127, 127, 127} % GPT-4o - 灰
    \definecolor{c4}{RGB}{255, 192, 0}   % GPT-5.2 - 黄
    \definecolor{c5}{RGB}{91, 155, 213}  % Gemini - 浅蓝
    \scalebox{0.82}{
    \begin{tikzpicture}
        \begin{axis}[
            width=\linewidth, % 宽度占满单栏
            height=7cm,       % 高度适中
            % --- 核心设置：横向点图 ---
            xmin=70, xmax=102, % 设置X轴范围，聚焦差异区域 (70%-100%)
            ymin=0.5, ymax=6.5, % Y轴范围
            % --- Y轴设置 (数据集名称) ---
            ytick={1,2,3,4,5,6},
            yticklabels={
                gsm8k,           % 放在最下面，因为差异最大
                CulturalBench-H,
                MATH-500,
                MathVista,
                CulturalBench-E,
                DocVQA           % 放在最上面，因为最稳定，奠定基调
            },
            yticklabel style={font=\small, align=right}, % Y轴标签样式
            y axis line style={opacity=0}, % 隐藏Y轴线，更干净
            % --- X轴设置 (准确率) ---
            xlabel={Accuracy (\%)},
            xticklabel style={font=\footnotesize, /pgf/number format/fixed},
            xmajorgrids=true, % 显示垂直网格辅助读数
            grid style={dashed, gray!30},
            % --- 图例设置 ---
            legend style={
                at={(0.5,1.05)}, % 放在图上方
                anchor=south,
                legend columns=5, % 一排展示
                draw=none, % 无边框
                font=\footnotesize
            },
            % --- 所有点的通用样式 ---
            only marks, % 只画点
            mark size=2.8pt, % 点的大小
            mark options={line width=0.8pt}, % 点的边缘
            % --- 修复 cycle list 语法 ---
            cycle list={
                {color=c1, mark=*, fill=c1!80},          % Claude - 实心圆
                {color=c2, mark=square*, fill=c2!80},    % Qwen - 实心方
                {color=c3, mark=triangle*, fill=c3!80},  % GPT-4o - 实心三角
                {color=c4, mark=diamond*, fill=c4!80},   % GPT-5.2 - 实心菱形
                {color=c5, mark=star, mark size=3.5pt, thick}, % Gemini - 星号(突出)
            }
        ]

        % --- 视觉核心：为每一行添加水平参考线 ---
        % 这条线让点看起来像是"穿在一条线上"，强调它们的聚集
        \draw[gray!20, line width=3pt, line cap=round] (axis cs:70,1) -- (axis cs:100,1);
        \draw[gray!20, line width=3pt, line cap=round] (axis cs:70,2) -- (axis cs:100,2);
        \draw[gray!20, line width=3pt, line cap=round] (axis cs:70,3) -- (axis cs:100,3);
        \draw[gray!20, line width=3pt, line cap=round] (axis cs:70,4) -- (axis cs:100,4);
        \draw[gray!20, line width=3pt, line cap=round] (axis cs:70,5) -- (axis cs:100,5);
        \draw[gray!20, line width=3pt, line cap=round] (axis cs:70,6) -- (axis cs:100,6);

        % --- 数据输入区域 ---
        % y坐标对应上面的 ytick labels 顺序。
        % 6=DocVQA, 5=CB-E, 4=MathVista, 3=MATH-500, 2=CB-H, 1=gsm8k

        % Claude 3.7
        \addplot+[color=c1, mark=*, fill=c1!80, mark size=2.8pt] coordinates {(99.66,6) (88.25,5) (85.45,4) (85.21,3) (81.80,2) (86.10,1)};
        \addlegendentry{Claude 3.7}

        % Qwen-VL-72B
        \addplot+[color=c2, mark=square*, fill=c2!80, mark size=2.8pt] coordinates {(99.44,6) (88.25,5) (85.88,4) (84.51,3) (82.21,2) (75.21,1)};
        \addlegendentry{Qwen 72B}

        % GPT-4o
        \addplot+[color=c3, mark=triangle*, fill=c3!80, mark size=2.8pt] coordinates {(99.44,6) (88.41,5) (86.85,4) (85.92,3) (82.21,2) (80.90,1)};
        \addlegendentry{GPT-4o}

        % GPT-5.2
        \addplot+[color=c4, mark=diamond*, fill=c4!80, mark size=2.8pt] coordinates {(99.55,6) (88.10,5) (86.85,4) (85.92,3) (81.97,2) (85.01,1)};
        \addlegendentry{GPT-5.2}

        % Gemini 3 Pro (星号突出)
        \addplot+[color=c5, mark=star, mark size=3.5pt, thick] coordinates {(99.44,6) (87.94,5) (85.78,4) (92.25,3) (82.21,2) (95.39,1)};
        \addlegendentry{Gemini 3 Pro}

        \end{axis}
    \end{tikzpicture}}
    \caption{Architectural Robustness Analysis. The tight clustering of colored markers along the horizontal tracks (especially in DocVQA, CulturalBench, and MathVista) visually demonstrates the high stability of the TwC paradigm across diverse MLLMs architectures. Notable outliers indicate model-specific strengths (e.g., Gemini on gsm8k) rather than method failure.}
    \label{fig:cross_model_dotplot}
\end{figure}
\vskip -8pt

Results are summarized in Figure~\ref{fig:cross_model_dotplot}. Results across different tasks show that TwC path $\mathrm{II}$ leads to largely consistent performance trends across models. On the DocVQA benchmark, all models maintain accuracy above 99.4\%, indicating that emphasizing key visual regions in comics, together with accompanying textual prompts, provides reliable auxiliary information. Notably, Gemini 3 Pro achieves relatively stronger performance on several tasks, reaching 95.3\% accuracy on GSM8K. Overall, comic panels function as a reusable intermediate representation that delivers stable performance gains across tasks and model configurations, demonstrating a certain degree of cross-model generalization.

\subsection{Efficiency Analysis of TwC and Think with Video}
% To demonstrate the practical value of the TwC paradigm compared to video-based reasoning, we select problems with strong dynamic attributes. We directly compare the reasoning chains of TwC against video reasoning chains (leveraging Sora-generated content) in terms of accuracy and resource consumption. Metrics include average total tokens, VRAM occupancy, and end-to-end latency.

To formalize the economic feasibility, we define the different visual signal generation cost function $C(\cdot)$. For video generation (Think with Video), the cost is time-dependent: $C_{video}(t) = \alpha \cdot t$, where $\alpha$ denotes the unit price per second. For our comic-based approach (TwC), the cost is image-dependent: $C_{comic} = \beta$, where $\beta$ represents the fixed cost of a single composite image.

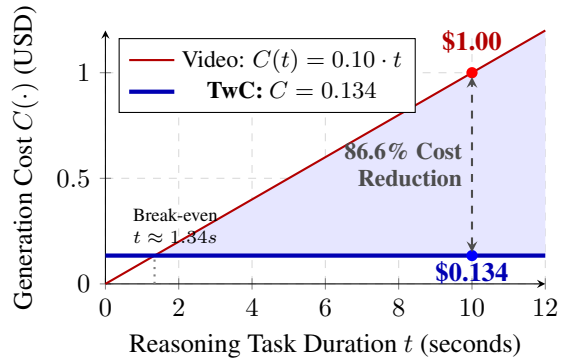
\begin{figure}[!ht]
\centering
\begin{tikzpicture}
\begin{axis}[
    width=0.9\linewidth,
    height=0.6\linewidth,
    xlabel={Reasoning Task Duration $t$ (seconds)},
    ylabel={Generation Cost $C(\cdot)$ (USD)},
    xmin=0, xmax=12,
    ymin=0, ymax=1.2,
    axis lines=left,
    grid=major,
    grid style={dashed, gray!30},
    legend pos=north west,
    legend style={font=\small},
    % 关键点：让坐标轴不从0紧贴，留点空隙好看
    enlargelimits=false,
    clip=false % 允许画在图外
]

% 1. 绘制 Video Cost 曲线 (y = 0.1x)
\addplot[
    domain=0:12, 
    samples=10, 
    color=red!70!black, 
    thick,
    name path=video
]
{0.1*x};
\addlegendentry{Video: $C(t) = 0.10 \cdot t$}

% 2. 绘制 Comic Cost 曲线 (y = 0.134)
\addplot[
    domain=0:12, 
    samples=10, 
    color=blue!70!black, 
    ultra thick,
    name path=comic
]
{0.134};
\addlegendentry{\textbf{TwC: $C = 0.134$}}

% 3. 填充"节省成本"的区域 (Efficiency Gain)
\addplot[blue!10] fill between[of=video and comic, soft clip={domain=1.34:12}];

% 4. 标注具体的 10s 实验点
% Video 点
\node[circle, fill=red, inner sep=1.5pt] at (axis cs:10, 1.0) {};
\node[red!70!black, anchor=south] at (axis cs:10, 1.05) {\textbf{\$1.00}};

% Comic 点
\node[circle, fill=blue, inner sep=1.5pt] at (axis cs:10, 0.134) {};
\node[blue!70!black, anchor=north] at (axis cs:10, 0.15) {\textbf{\$0.134}};

% 5. 画出差距箭头
\draw[<->, >=stealth, thick, dashed, black!70] (axis cs:10, 0.15) -- (axis cs:10, 0.98) 
    node[midway, left, align=right, font=\footnotesize] {\textbf{86.6\% Cost}\\\textbf{Reduction}};

% 6. (可选) 标注盈亏平衡点 Break-even point
\draw[dotted, thick, gray] (axis cs:1.34, 0) -- (axis cs:1.34, 0.134);
\node[font=\scriptsize, align=center, anchor=south west] at (axis cs:0.50, 0.15) {Break-even\\$t \approx 1.34s$};

\end{axis}
\end{tikzpicture}
\caption{Comparing the image generation cost models. While video generation cost ($C_{video}$) scales linearly with task duration due to temporal redundancy, TwC maintains a low, constant cost ($C_{comic}$) regardless of the event's temporal length. The shaded area represents the economic advantage of our approach.}
\label{fig:cost_scaling}
\end{figure}
\vskip -7pt

% Adopting the standard industrial pricing ($\alpha=\$0.10/s$~\footnote{https://openai.com/api/pricing/}, $\beta=\$0.134/img$~\footnote{https://ai.google.dev/gemini-api/docs/pricing}), a standard 10-second dynamic reasoning task incurs a cost of $1.00$ via video generation but only $0.134$ via TwC. 

% Adopting standard industrial pricing ($\alpha=$0.10/\text{s}$, $\beta=$0.134/\text{img}$), a 10-second dynamic reasoning task under the \textit{Thinking with Video} setting (consistent with prior work) costs $1.00 via video generation, compared to only $0.134 with TwC. This corresponds to a cost compression ratio of $\frac{C_{comic}}{C_{video}} \approx 13.4\%$, i.e., an 86.6\% reduction in media generation cost for a typical reasoning instance. Notably, the two cost functions intersect at a break-even point of $t \approx 1.34,\mathrm{s}$, beyond which video-based reasoning becomes strictly more expensive. These results demonstrate that TwC achieves an reduction in computational overhead without compromising the reasoning accuracy.

Adopting standard industrial pricing ($\alpha=\$0.10/\text{s}$~\footnote{https://openai.com/api/pricing/}, $\beta=\$0.134/\text{img}$~\footnote{https://ai.google.dev/gemini-api/docs/pricing}), a 10-second dynamic reasoning task under the \textit{Thinking with Video~\cite{tong2025thinking}} setting (consistent with prior work) costs \$1.00 via video generation, compared to only \$0.134 with TwC. This corresponds to a cost compression ratio of $\frac{C_{comic}}{C_{video}} \approx 13.4\%$, i.e., an 86.6\% reduction in media generation cost for a typical reasoning instance. Notably, the two cost functions intersect at a break-even point of $t \approx 1.34\,\mathrm{s}$, beyond which video-based reasoning becomes strictly more expensive. These results demonstrate that TwC achieves a reduction in computational overhead 
without compromising reasoning accuracy. We theoretically analyze in Appendix~\ref{budegt_comics_video_appendix} why comics are more budget-efficient than videos.

% This yields a cost compression ratio of $\frac{C_{comic}}{C_{video}} \approx 13.4\%$. 

% As shown in the Pareto scatter plot in Figure~\ref{fig:efficiency_api}, TwC maintains approximately 92\% of the accuracy of video reasoning while incurring only 15\% of the computational cost. The results intuitively reveal the severe information redundancy inherent in the video modality during reasoning processes. We conclude that comics, as a temporal compression medium, offer an industrial-grade solution with the highest cost-performance ratio for long-sequence logic tasks.

\section{Conclusion}
% \section{Conclusion}
We introduce \textit{Thinking with Comics}, a multimodal reasoning paradigm that uses multi-panel comics as an efficient intermediate representation for temporal and multi-step reasoning. TwC improves reasoning performance while avoiding video-generation overhead, with analyses highlighting the roles of narrative structure and embedded text, pointing to future directions in controllability, faithfulness, and evaluation.

\section*{Impact Statement}
This paper proposes Thinking with Comics, an efficient multimodal reasoning paradigm that uses comics as an intermediate representation between images and videos. By reducing redundancy and computational cost while preserving temporal and narrative structure, the approach improves the efficiency and practicality of multimodal reasoning systems for long-context and temporal reasoning tasks. We do not foresee immediate harmful applications; nevertheless, future work should consider the influence of narrative style and cultural conventions in comics to ensure robust and fair deployment across diverse settings.

% Authors are \textbf{required} to include a statement of the potential broader
% impact of their work, including its ethical aspects and future societal
% consequences. This statement should be in an unnumbered section at the end of
% the paper (co-located with Acknowledgements -- the two may appear in either
% order, but both must be before References), and does not count toward the paper
% page limit. In many cases, where the ethical impacts and expected societal
% implications are those that are well established when advancing the field of
% Machine Learning, substantial discussion is not required, and a simple
% statement such as the following will suffice:

% This paper presents work whose goal is to advance the field of Machine
% Learning. There are many potential societal consequences of our work, none
% which we feel must be specifically highlighted here.

% The above statement can be used verbatim in such cases, but we encourage
% authors to think about whether there is content which does warrant further
% discussion, as this statement will be apparent if the paper is later flagged
% for ethics review.

% In the unusual situation where you want a paper to appear in the
% references without citing it in the main text, use \nocite
% \nocite{langley00}

\bibliography{arxiv/example_paper}

@article{wei2022chain,
  title={Chain-of-thought prompting elicits reasoning in large language models},
  author={Wei, Jason and Wang, Xuezhi and Schuurmans, Dale and Bosma, Maarten and Xia, Fei and Chi, Ed and Le, Quoc V and Zhou, Denny and others},
  journal={Advances in neural information processing systems},
  volume={35},
  pages={24824--24837},
  year={2022}
}

@article{cheng2025visual,
  title={Visual thoughts: A unified perspective of understanding multimodal chain-of-thought},
  author={Cheng, Zihui and Chen, Qiguang and Xu, Xiao and Wang, Jiaqi and Wang, Weiyun and Fei, Hao and Wang, Yidong and Wang, Alex Jinpeng and Chen, Zhi and Che, Wanxiang and others},
  journal={arXiv preprint arXiv:2505.15510},
  year={2025}
}

@article{tong2025thinking,
  title={Thinking with video: Video generation as a promising multimodal reasoning paradigm},
  author={Tong, Jingqi and Mou, Yurong and Li, Hangcheng and Li, Mingzhe and Yang, Yongzhuo and Zhang, Ming and Chen, Qiguang and Liang, Tianyi and Hu, Xiaomeng and Zheng, Yining and others},
  journal={arXiv preprint arXiv:2511.04570},
  year={2025}
}

@article{hurst2024gpt,
  title={Gpt-4o system card},
  author={Hurst, Aaron and Lerer, Adam and Goucher, Adam P and Perelman, Adam and Ramesh, Aditya and Clark, Aidan and Ostrow, AJ and Welihinda, Akila and Hayes, Alan and Radford, Alec and others},
  journal={arXiv preprint arXiv:2410.21276},
  year={2024}
}

@article{kojima2022large,
  title={Large language models are zero-shot reasoners},
  author={Kojima, Takeshi and Gu, Shixiang Shane and Reid, Machel and Matsuo, Yutaka and Iwasawa, Yusuke},
  journal={Advances in neural information processing systems},
  volume={35},
  pages={22199--22213},
  year={2022}
}

@article{wang2022self,
  title={Self-consistency improves chain of thought reasoning in language models},
  author={Wang, Xuezhi and Wei, Jason and Schuurmans, Dale and Le, Quoc and Chi, Ed and Narang, Sharan and Chowdhery, Aakanksha and Zhou, Denny},
  journal={arXiv preprint arXiv:2203.11171},
  year={2022}
}

@inproceedings{huang2023towards,
  title={Towards reasoning in large language models: A survey},
  author={Huang, Jie and Chang, Kevin Chen-Chuan},
  booktitle={Findings of the association for computational linguistics: ACL 2023},
  pages={1049--1065},
  year={2023}
}

@article{zhang2023multimodal,
  title={Multimodal chain-of-thought reasoning in language models},
  author={Zhang, Zhuosheng and Zhang, Aston and Li, Mu and Zhao, Hai and Karypis, George and Smola, Alex},
  journal={arXiv preprint arXiv:2302.00923},
  year={2023}
}

@article{zheng2023ddcot,
  title={Ddcot: Duty-distinct chain-of-thought prompting for multimodal reasoning in language models},
  author={Zheng, Ge and Yang, Bin and Tang, Jiajin and Zhou, Hong-Yu and Yang, Sibei},
  journal={Advances in Neural Information Processing Systems},
  volume={36},
  pages={5168--5191},
  year={2023}
}

@inproceedings{mitra2024compositional,
  title={Compositional chain-of-thought prompting for large multimodal models},
  author={Mitra, Chancharik and Huang, Brandon and Darrell, Trevor and Herzig, Roei},
  booktitle={Proceedings of the IEEE/CVF Conference on Computer Vision and Pattern Recognition},
  pages={14420--14431},
  year={2024}
}

@inproceedings{gao2024cantor,
  title={Cantor: Inspiring multimodal chain-of-thought of mllm},
  author={Gao, Timin and Chen, Peixian and Zhang, Mengdan and Fu, Chaoyou and Shen, Yunhang and Zhang, Yan and Zhang, Shengchuan and Zheng, Xiawu and Sun, Xing and Cao, Liujuan and others},
  booktitle={Proceedings of the 32nd ACM International Conference on Multimedia},
  pages={9096--9105},
  year={2024}
}

@article{shao2024deepseekmath,
  title={Deepseekmath: Pushing the limits of mathematical reasoning in open language models},
  author={Shao, Zhihong and Wang, Peiyi and Zhu, Qihao and Xu, Runxin and Song, Junxiao and Bi, Xiao and Zhang, Haowei and Zhang, Mingchuan and Li, YK and Wu, Yang and others},
  journal={arXiv preprint arXiv:2402.03300},
  year={2024}
}

@article{guo2025deepseek,
  title={Deepseek-r1: Incentivizing reasoning capability in llms via reinforcement learning},
  author={Guo, Daya and Yang, Dejian and Zhang, Haowei and Song, Junxiao and Zhang, Ruoyu and Xu, Runxin and Zhu, Qihao and Ma, Shirong and Wang, Peiyi and Bi, Xiao and others},
  journal={arXiv preprint arXiv:2501.12948},
  year={2025}
}

@article{liu2025visual,
  title={Visual-rft: Visual reinforcement fine-tuning},
  author={Liu, Ziyu and Sun, Zeyi and Zang, Yuhang and Dong, Xiaoyi and Cao, Yuhang and Duan, Haodong and Lin, Dahua and Wang, Jiaqi},
  journal={arXiv preprint arXiv:2503.01785},
  year={2025}
}

@inproceedings{xu2025llava,
  title={Llava-cot: Let vision language models reason step-by-step},
  author={Xu, Guowei and Jin, Peng and Wu, Ziang and Li, Hao and Song, Yibing and Sun, Lichao and Yuan, Li},
  booktitle={Proceedings of the IEEE/CVF International Conference on Computer Vision},
  pages={2087--2098},
  year={2025}
}

@inproceedings{dhuliawala2024chain,
  title={Chain-of-verification reduces hallucination in large language models},
  author={Dhuliawala, Shehzaad and Komeili, Mojtaba and Xu, Jing and Raileanu, Roberta and Li, Xian and Celikyilmaz, Asli and Weston, Jason},
  booktitle={Findings of the association for computational linguistics: ACL 2024},
  pages={3563--3578},
  year={2024}
}

@article{ho2020denoising,
  title={Denoising diffusion probabilistic models},
  author={Ho, Jonathan and Jain, Ajay and Abbeel, Pieter},
  journal={Advances in neural information processing systems},
  volume={33},
  pages={6840--6851},
  year={2020}
}

@inproceedings{rombach2022high,
  title={High-resolution image synthesis with latent diffusion models},
  author={Rombach, Robin and Blattmann, Andreas and Lorenz, Dominik and Esser, Patrick and Ommer, Bj{\"o}rn},
  booktitle={Proceedings of the IEEE/CVF conference on computer vision and pattern recognition},
  pages={10684--10695},
  year={2022}
}

@article{betker2023improving,
  title={Improving image generation with better captions},
  author={Betker, James and Goh, Gabriel and Jing, Li and Brooks, Tim and Wang, Jianfeng and Li, Linjie and Ouyang, Long and Zhuang, Juntang and Lee, Joyce and Guo, Yufei and others},
  journal={Computer Science. https://cdn. openai. com/papers/dall-e-3. pdf},
  volume={2},
  number={3},
  pages={8},
  year={2023}
}

@inproceedings{chen2025make,
  title={Make imagination clearer! stable diffusion-based visual imagination for multimodal machine translation},
  author={Chen, Andong and Song, Yuchen and Chen, Kehai and Bai, Xuefeng and Yang, Muyun and Nie, Liqiang and Liu, Jie and Zhao, Tiejun and Zhang, Min},
  booktitle={Proceedings of the 63rd Annual Meeting of the Association for Computational Linguistics (Volume 1: Long Papers)},
  pages={26567--26583},
  year={2025}
}

@inproceedings{OpenAIOA,
  title={OpenAI o3 and o4-mini System Card},
  author={},
  url={https://api.semanticscholar.org/CorpusID:278283461}
}

@article{singh2025openai,
  title={Openai gpt-5 system card},
  author={Singh, Aaditya and Fry, Adam and Perelman, Adam and Tart, Adam and Ganesh, Adi and El-Kishky, Ahmed and McLaughlin, Aidan and Low, Aiden and Ostrow, AJ and Ananthram, Akhila and others},
  journal={arXiv preprint arXiv:2601.03267},
  year={2025}
}

@misc{sora2,
	title={Sora 2 is here},
	author={OpenAI},
	howpublished={\url{https://openai.com/index/sora-2/}}, 
	year={2025}
}

@misc{veo3,
	title={Gemini AI video generator powered by Veo 3.1},
	author={Google},
	howpublished={\url{https://gemini.google/overview/video-generation/}}, 
	year={2025}
}

@inproceedings{lightman2023let,
  title={Let's verify step by step},
  author={Lightman, Hunter and Kosaraju, Vineet and Burda, Yuri and Edwards, Harrison and Baker, Bowen and Lee, Teddy and Leike, Jan and Schulman, John and Sutskever, Ilya and Cobbe, Karl},
  booktitle={The Twelfth International Conference on Learning Representations},
  year={2023}
}

@article{cobbe2021training,
  title={Training verifiers to solve math word problems},
  author={Cobbe, Karl and Kosaraju, Vineet and Bavarian, Mohammad and Chen, Mark and Jun, Heewoo and Kaiser, Lukasz and Plappert, Matthias and Tworek, Jerry and Hilton, Jacob and Nakano, Reiichiro and others},
  journal={arXiv preprint arXiv:2110.14168},
  year={2021}
}

@article{lu2023mathvista,
  title={Mathvista: Evaluating mathematical reasoning of foundation models in visual contexts},
  author={Lu, Pan and Bansal, Hritik and Xia, Tony and Liu, Jiacheng and Li, Chunyuan and Hajishirzi, Hannaneh and Cheng, Hao and Chang, Kai-Wei and Galley, Michel and Gao, Jianfeng},
  journal={arXiv preprint arXiv:2310.02255},
  year={2023}
}

@inproceedings{guerin2013ebdtheque,
  title={eBDtheque: a representative database of comics},
  author={Gu{\'e}rin, Cl{\'e}ment and Rigaud, Christophe and Mercier, Antoine and Ammar-Boudjelal, Farid and Bertet, Karell and Bouju, Alain and Burie, Jean-Christophe and Louis, Georges and Ogier, Jean-Marc and Revel, Arnaud},
  booktitle={2013 12th International Conference on Document Analysis and Recognition},
  pages={1145--1149},
  year={2013},
  organization={IEEE}
}

@inproceedings{mathew2021docvqa,
  title={Docvqa: A dataset for vqa on document images},
  author={Mathew, Minesh and Karatzas, Dimosthenis and Jawahar, CV},
  booktitle={Proceedings of the IEEE/CVF winter conference on applications of computer vision},
  pages={2200--2209},
  year={2021}
}

@article{chiu2024culturalbench,
  title={CulturalBench: A Robust, Diverse, and Challenging Cultural Benchmark by Human-AI CulturalTeaming},
  author={Chiu, Yu Ying and Jiang, Liwei and Lin, Bill Yuchen and Park, Chan Young and Li, Shuyue Stella and Ravi, Sahithya and Bhatia, Mehar and Antoniak, Maria and Tsvetkov, Yulia and Shwartz, Vered and others},
  journal={arXiv preprint arXiv:2410.02677},
  year={2024}
}

@misc{ClaudeSonnet45,
	title={Introducing Claude Sonnet 4.5},
	author={Anthropic},
	howpublished={\url{https://www.anthropic.com/news/claude-sonnet-4-5}}, 
	year={2025}
}

@misc{Gemini3Pro,
	title={Gemini 3 Pro},
	author={{Google DeepMind}},
	howpublished={\url{https://deepmind.google/models/gemini/pro/}}, 
	year={2025}
}

@article{yang2025qwen3,
  title={Qwen3 technical report},
  author={Yang, An and Li, Anfeng and Yang, Baosong and Zhang, Beichen and Hui, Binyuan and Zheng, Bo and Yu, Bowen and Gao, Chang and Huang, Chengen and Lv, Chenxu and others},
  journal={arXiv preprint arXiv:2505.09388},
  year={2025}
}

@article{dong2023dreamllm,
  title={Dreamllm: Synergistic multimodal comprehension and creation},
  author={Dong, Runpei and Han, Chunrui and Peng, Yuang and Qi, Zekun and Ge, Zheng and Yang, Jinrong and Zhao, Liang and Sun, Jianjian and Zhou, Hongyu and Wei, Haoran and others},
  journal={arXiv preprint arXiv:2309.11499},
  year={2023}
}

@inproceedings{besta2024graph,
  title={Graph of thoughts: Solving elaborate problems with large language models},
  author={Besta, Maciej and Blach, Nils and Kubicek, Ales and Gerstenberger, Robert and Podstawski, Michal and Gianinazzi, Lukas and Gajda, Joanna and Lehmann, Tomasz and Niewiadomski, Hubert and Nyczyk, Piotr and others},
  booktitle={Proceedings of the AAAI conference on artificial intelligence},
  volume={38},
  number={16},
  pages={17682--17690},
  year={2024}
}

@article{yao2023tree,
  title={Tree of thoughts: Deliberate problem solving with large language models},
  author={Yao, Shunyu and Yu, Dian and Zhao, Jeffrey and Shafran, Izhak and Griffiths, Tom and Cao, Yuan and Narasimhan, Karthik},
  journal={Advances in neural information processing systems},
  volume={36},
  pages={11809--11822},
  year={2023}
}

@article{wang2025multimodal,
  title={Multimodal chain-of-thought reasoning: A comprehensive survey},
  author={Wang, Yaoting and Wu, Shengqiong and Zhang, Yuecheng and Yan, Shuicheng and Liu, Ziwei and Luo, Jiebo and Fei, Hao},
  journal={arXiv preprint arXiv:2503.12605},
  year={2025}
}

@article{li2025imagine,
  title={Imagine while reasoning in space: Multimodal visualization-of-thought},
  author={Li, Chengzu and Wu, Wenshan and Zhang, Huanyu and Xia, Yan and Mao, Shaoguang and Dong, Li and Vuli{\'c}, Ivan and Wei, Furu},
  journal={arXiv preprint arXiv:2501.07542},
  year={2025}
}

@article{hu2024visual,
  title={Visual sketchpad: Sketching as a visual chain of thought for multimodal language models},
  author={Hu, Yushi and Shi, Weijia and Fu, Xingyu and Roth, Dan and Ostendorf, Mari and Zettlemoyer, Luke and Smith, Noah A and Krishna, Ranjay},
  journal={Advances in Neural Information Processing Systems},
  volume={37},
  pages={139348--139379},
  year={2024}
}

@inproceedings{augereau2017overview,
  title={An overview of comics research in computer science},
  author={Augereau, Olivier and Iwata, Motoi and Kise, Koichi},
  booktitle={2017 14th IAPR International Conference on Document Analysis and Recognition (ICDAR)},
  volume={3},
  pages={54--59},
  year={2017},
  organization={IEEE}
}
\bibliographystyle{arxiv/icml2026}

%%%%%%%%%%%%%%%%%%%%%%%%%%%%%%%%%%%%%%%%%%%%%%%%%%%%%%%%%%%%%%%%%%%%%%%%%%%%%%%
%%%%%%%%%%%%%%%%%%%%%%%%%%%%%%%%%%%%%%%%%%%%%%%%%%%%%%%%%%%%%%%%%%%%%%%%%%%%%%%
% APPENDIX
%%%%%%%%%%%%%%%%%%%%%%%%%%%%%%%%%%%%%%%%%%%%%%%%%%%%%%%%%%%%%%%%%%%%%%%%%%%%%%%
%%%%%%%%%%%%%%%%%%%%%%%%%%%%%%%%%%%%%%%%%%%%%%%%%%%%%%%%%%%%%%%%%%%%%%%%%%%%%%%
\newpage
\appendix
\onecolumn

\section{Empirical Analysis: Why Comics Are a Privileged Visual Reasoning Medium}
\label{appendix_photo_comics}

Building on the theoretical analysis in Appendix~\ref{start_analysis}, this section empirically evaluates the advantages of comics as a visual reasoning medium. Specifically, we investigate (i) the structural stability of comic-based multi-panel generation, and (ii) the benefits of treating comics as a global structure compared to incremental visual reasoning.

\subsection{Prompt-Induced Structural Stability in Multi-Panel Visual Generation}
\label{appendix_prompt}
This experiment examines whether comics, compared to non-comic visual styles, more naturally and stably support multi-panel generation. This setting is motivated by our theoretical analysis in Appendix~\ref{appendix_domain_shift}.
We design two controlled prompt settings. In the Comic condition, the model is instructed to “draw a four-panel comic to solve the problem.” In the Non-Comic condition, the model is instructed to “draw a four-step visual storyboard in a realistic style,” with the number of panels explicitly constrained to match the comic setting. Except for the presence of the word ``comic'', all other prompt components and decoding parameters are kept identical.

For evaluation, we sample 20 instances each from MATH-500 and MathVista. The generated images are answered by Gemini-3 Pro. We evaluate (i) the success rate of generating the required number of panels, and (ii) answer accuracy.
\begin{table}[!ht]
\centering
\label{tab:prompt_stability}
\caption{Comparison of structural stability and reasoning accuracy between Comic and Non-Comic prompts on MATH-500 and MathVista.}
\begin{tabular}{@{}llccc@{}}
\toprule
\textbf{Metric} & \textbf{Dataset} & \textbf{Comic} & \textbf{Non-Comic} & \textbf{Improvement} \\ \midrule
\multirow{2}{*}{\begin{tabular}[c]{@{}l@{}}Layout Success Rate (\%)\\ (Panel Consistency)\end{tabular}} & MATH-500 & \textbf{95.0} & 70.0 & +25.0 \\
 & MathVista & \textbf{90.0} & 65.0 & +25.0 \\ \midrule
\multirow{2}{*}{\begin{tabular}[c]{@{}l@{}}Reasoning Accuracy (\%)\\ \end{tabular}} & MATH-500 & \textbf{75.0} & 60.0 & +15.0 \\
 & MathVista & \textbf{70.0} & 55.0 & +15.0 \\ \bottomrule
\end{tabular}
\label{multi-plane}
\end{table}

As shown in Table~\ref{multi-plane}, comic prompts consistently induce structurally complete multi-panel layouts across tasks, whereas Non-Comic instructions more frequently suffer from layout collapse or unintended merging of multiple steps, failing to reliably satisfy the step-wise generation constraint. The inherent panel-based structure of comics provides a strong structural prior, aligning multi-step visual reasoning with chain-of-thought in the visual domain, and thereby improving the stability and overall performance of multimodal reasoning. These observations provide empirical support for our domain-shift analysis, suggesting that the comic format offers a natural and robust scaffold for multi-panel generation that is difficult to reproduce with ad-hoc non-comic visual styles.

\subsection{Structural Coherence in Global vs. Incremental Visual Reasoning}
\label{appendix_reasoing}

This experiment compares Global Comic generation and Incremental image chaining for multi-step visual reasoning. This comparison is motivated by our theoretical analysis in Appendix~\ref{appendix_multi_plane_analysis}.
The former generates a complete multi-panel comic in a single pass, while the latter produces panels sequentially conditioned on previous outputs, with an identical number of panels in both settings. We evaluate on 20 samples each from MATH-500 and MathVista using human judgments on cross-panel logical continuity, state transitions, and textual quality (Appendix~\ref{human_evaluation}).

\begin{table}[!ht]
\centering
\label{tab:human_eval_generation}
% \caption{Human evaluation results comparing Global Comic and Incremental generation. Scores are on a 1-5 scale (higher is better). Global generation shows significant superiority in structural coherence.}
\caption{Human evaluation results comparing Global and Incremental generation. We evaluate Accuracy (ACC) and three structural metrics (1-5 scale): \textbf{Logic} (reasoning flow), \textbf{State} (consistency between panels), and \textbf{Quality} (visual-textual fidelity). Global generation shows significant superiority in both objective performance and structural coherence.}
\begin{tabular}{@{}llcccc@{}}
\toprule
\textbf{Benchmark} & \textbf{Method} & \textbf{ACC (\%)} $\uparrow$ & \textbf{Logic} $\uparrow$ & \textbf{State} $\uparrow$ & \textbf{Quality} $\uparrow$ \\ \midrule
\multirow{2}{*}{MATH-500} & Incremental & 80.0 & 4.17 & 3.72 & 3.58 \\
 & Global (Ours) & \textbf{95.0} & \textbf{4.86} & \textbf{4.67} & \textbf{4.61} \\ \midrule
\multirow{2}{*}{MathVista} & Incremental & 50.0 & 3.50 & 3.50 & 3.40 \\
 & Global (Ours) & \textbf{85.0} & \textbf{4.47} & \textbf{4.45} & \textbf{4.58} \\ \midrule
\multirow{2}{*}{Average} & Incremental & 65.0 & 3.83 & 3.61 & 3.49 \\
 & Global (Ours) & \textbf{90.0} & \textbf{4.67} & \textbf{4.56} & \textbf{4.59} \\ \bottomrule
\end{tabular}
\label{global_incremental}
\end{table}

% \begin{tabular}{@{}llcccc@{}}
% \toprule
% \textbf{Benchmark} & \textbf{Method} & \textbf{ACC (\%)} $\uparrow$ & \textbf{Logic} $\uparrow$ & \textbf{State} $\uparrow$ & \textbf{Quality} $\uparrow$ \\ \midrule
% \multirow{2}{*}{MATH-500} & Incremental & 80.0 & 4.17 & 3.72 & 3.58 \\
%  & Global (Ours) & \textbf{95.0} & \textbf{4.86} & \textbf{4.67} & \textbf{4.61} \\ \midrule
% \multirow{2}{*}{MathVista} & Incremental & 50.0 & 3.50 & 3.50 & 3.40 \\
%  & Global (Ours) & \textbf{85.0} & \textbf{4.47} & \textbf{4.45} & \textbf{4.58} \\ \midrule
% \multirow{2}{*}{Average} & Incremental & 65.0 & 3.83 & 3.61 & 3.49 \\
%  & Global (Ours) & \textbf{90.0} & \textbf{4.67} & \textbf{4.56} & \textbf{4.59} \\ \bottomrule
% \end{tabular}

\begin{figure}[!ht]
\centering
\includegraphics[width=\linewidth]{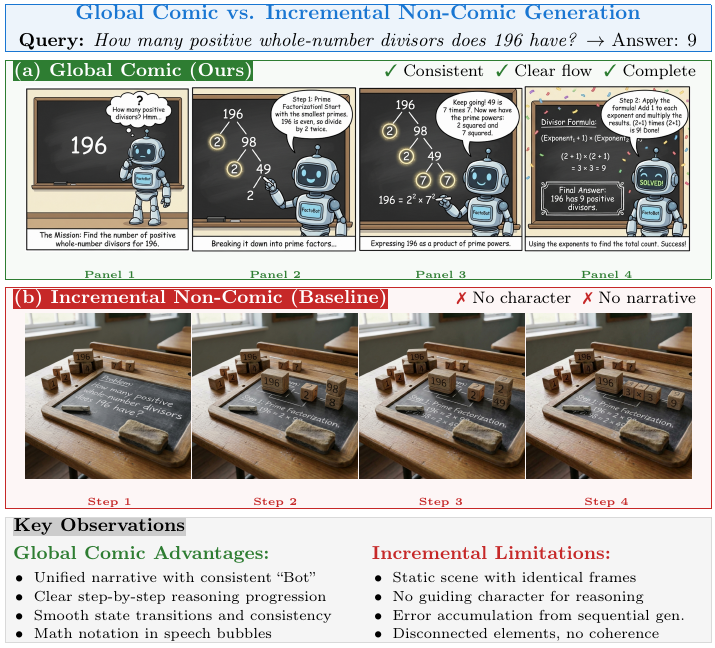} 
\caption{Qualitative comparison between (a) Global Comic (Ours) and (b) Incremental Non-Comic generation for a mathematical reasoning task (finding divisors of 196). Global generation maintains a consistent character (FactoBot) and smooth logical flow, whereas the incremental baseline exhibits static scenes and lacks narrative coherence.}
\label{fig:qualitative_comparison}
\end{figure}

As shown in Table~\ref{global_incremental} and a bad case (Figure~\ref{fig:qualitative_comparison}), results show that global generation yields significantly stronger cross-panel coherence, with more stable entity representations and smoother reasoning progression, whereas incremental generation suffers from error accumulation. This suggests that treating comics as a holistic structured representation is crucial for preserving multi-step reasoning quality. These findings empirically support our claim that global structural planning, rather than stepwise local generation, is essential for maintaining coherent multi-step reasoning trajectories in the visual domain.

\section{Theoretical Justification of Comics as a High-Quality and Efficient Visual Reasoning Medium}
\label{start_analysis}

\subsection{Representation and Utility}
Let $q$ denote a question, $a$ the ground-truth answer, and $z$ an intermediate representation generated by a visual generator $G_\theta$.
Under Path~II, the final prediction is $\hat a = F_\phi(q, z)$, consistent with Eq.~(4--6) in the main paper.

We characterize an intermediate representation $z$ by two orthogonal criteria:
\textbf{(i) generation fidelity} (how well $G_\theta$ can produce $z$), and
\textbf{(ii) task sufficiency} (how informative $z$ is for predicting $a$).

\paragraph{Information-efficiency.}
We define the \emph{information-efficiency} of $z$ for task solving as
\begin{equation}
\eta(z) \triangleq \frac{I(a; z \mid q)}{C(z)},
\end{equation}
where $I(\cdot;\cdot\mid\cdot)$ is conditional mutual information and $C(z)$ is the media generation cost.
Our main paper already instantiates $C(\cdot)$ for video and comics (constant per image vs.\ linear per second), providing an empirical cost rationale.

\subsection{Comics Outperform Single Images Due to Temporal Structure and Textual Anchoring}
\label{appendix_multi_plane_analysis}
A single image $x$ is typically a snapshot of an underlying latent trajectory $s_{1:T}$ (temporal/causal process).
If the answer $a$ depends on multi-step temporal or causal relations in $s_{1:T}$, then any snapshot $x = h(s_t)$ may discard relevant states.
Formally, whenever $a$ is not conditionally independent of the latent trajectory given a snapshot, i.e.,
\begin{equation}
I(a; s_{1:T} \mid q, x) > 0,
\end{equation}
we have a strict information gap:

% SYC MOD BEGIN
% Reason: Misuse of the formula:$I(X;Z) + I(X;Y|Z) = I(X;Y,Z)$
\begin{equation}
    I(a;s_{1:T} \mid q) = I(a;s_{1:T}, q) - I(a\mid q) > I(a;x,q) - I(a\mid q) = I(a;x\mid q)
\end{equation}
%\begin{equation}
%I(a; s_{1:T} \mid q) = I(a; x \mid q) + I(a; s_{1:T} \mid q, x) \;>\; I(a; x \mid q).
%\end{equation}
% SYC MOD END

Comics represent a structured summary $z_{\text{comic}} = (c_{1:K}, \tau)$ consisting of $K$ panels $c_{1:K}$ (selected intermediate states) and embedded text $\tau$ (bubbles/narration).
By the chain rule,
\begin{equation}
I(a; z_{\text{comic}} \mid q)
= I(a; c_{1:K} \mid q) + I(a; \tau \mid q, c_{1:K}),
\end{equation}
where the second term is \emph{non-negative} and captures the additional semantic anchoring channel.
Therefore, comics can strictly dominate pure-visual sequences whenever textual anchoring carries answer-relevant cues:
\begin{equation}
I(a; z_{\text{comic}} \mid q) \;\ge\; I(a; c_{1:K} \mid q),
\quad \text{and if } I(a;\tau\mid q,c_{1:K})>0 \text{ then the inequality is strict.}
\end{equation}
This matches our ablation that adding bubbles/narration improves robustness and accuracy.

\subsection{Comics Are More Efficient Than Videos Under a Budget}
\label{budegt_comics_video_appendix}
Let a video be $v = (x_1,\dots,x_T)$ with $T$ frames.
By the chain rule,
\begin{equation}
I(a; v \mid q) = \sum_{t=1}^{T} I(a; x_t \mid q, x_{<t}).
\end{equation}
In realistic videos, consecutive frames are highly correlated, hence $I(a; x_t \mid q, x_{<t})$ quickly diminishes as $t$ grows (temporal redundancy).
Thus, $I(a; v \mid q)$ grows \emph{sublinearly} with $T$ while video cost grows \emph{linearly} with $T$ (or duration).
Consequently, the efficiency $\eta(v) = I(a;v\mid q)/C(v)$ decreases with longer videos once redundancy dominates.

Comics can be seen as selecting $K \ll T$ \emph{key states} (panels) from the latent trajectory to maximize task-relevant information:
\begin{equation}
c_{1:K} \approx \arg\max_{S \subseteq \{1,\dots,T\},\, |S|=K} I\!\left(a; x_S \mid q\right).
\end{equation}
When the set function $f(S)=I(a;x_S\mid q)$ is approximately submodular (a standard diminishing-returns property for information measures),
greedy selection achieves a $(1-1/e)$ approximation to the optimal subset.
Hence, with far fewer visual tokens, comics retain most of the answer-relevant information while avoiding redundant frames,
leading to higher $\eta(z_{\text{comic}})$ than $\eta(v)$ at the same budget.
This aligns with our observed panel-scaling curve where accuracy saturates around $K\in[4,6]$.

\subsection{Why Comics Generate Better than Synthetic Sequential Images: A Domain-Shift Bound}
\label{appendix_domain_shift}
We now justify the claim that comics (a real, widely observed visual genre) are generated with higher fidelity than ad-hoc
``synthetic sequential images with logical relations'' that do not correspond to a well-established visual manifold.

Let $P_{\text{train}}$ be the (unknown) effective training distribution of the image generator.
Let $P_{\text{comic}}$ be the target distribution of real comics, and $P_{\text{syn}}$ the distribution of synthetic sequential images.
Consider a perceptual fidelity loss $\mathcal{L}(x)$ (e.g., measuring artifacts, inconsistency, or poor alignment with prompts).
A standard domain adaptation bound (Ben-David type) implies that for any hypothesis class induced by the generator,
\begin{equation}
\mathbb{E}_{x\sim P_{\text{target}}}[\mathcal{L}(x)]
\;\le\;
\mathbb{E}_{x\sim P_{\text{train}}}[\mathcal{L}(x)]
+ \mathrm{Div}(P_{\text{train}}, P_{\text{target}}) + \lambda,
\end{equation}
where $\mathrm{Div}(\cdot,\cdot)$ is a distribution divergence (e.g., $\mathcal{H}\Delta\mathcal{H}$-divergence or an IPM),
and $\lambda$ is the irreducible joint error term.
If comics are a \emph{real, frequent} genre, then $P_{\text{comic}}$ is closer to $P_{\text{train}}$ than an ad-hoc synthetic style:
\begin{equation}
\mathrm{Div}(P_{\text{train}}, P_{\text{comic}})
\;<\;
\mathrm{Div}(P_{\text{train}}, P_{\text{syn}}).
\end{equation}
Therefore, the expected fidelity loss is lower for comics:
\begin{equation}
\mathbb{E}_{x\sim P_{\text{comic}}}[\mathcal{L}(x)]
\;<\;
\mathbb{E}_{x\sim P_{\text{syn}}}[\mathcal{L}(x)],
\end{equation}
i.e., the generator produces higher-quality outputs in the comic domain than in a less natural, distribution-shifted synthetic domain.

% \paragraph{Takeaway.}
Comics simultaneously (i) reduce task uncertainty via structured temporal panels and textual anchoring, (ii) avoid the redundancy and high cost of video, and (iii) achieve higher generation fidelity due to smaller domain shift. Together, these provide a principled justification for Thinking with Comics as a high-density intermediate reasoning representation.

\section{Answer Extraction Protocol for Thinking with Comics}
\label{exact_final_answer_appendix}

\subsection{Path I: End-to-End Comic Reasoning.}
In Path~I, the model generates a multi-panel comic as the complete reasoning resut, where the final answer is visually embedded in the comic, typically appearing in the last panel as explicit text or a highlighted result. We perform answer extraction using \textbf{GPT-5.2} as an external answer reader (The detail of prompt is in Appendix~\ref{exact_answer_mllm}). The extractor is provided with the generated comic panels together with the original question, and is instructed to identify the final answer depicted in the comic.
The extracted answer is matched against the ground-truth label to compute ACC.

\paragraph{Human Verification for Path I.}
To validate the reliability of model-based answer extraction, we randomly sample \textbf{20\%} of the evaluation instances for manual inspection. For each sampled instance, a human annotator independently reads the answer from the comic and compares it with the answer extracted by GPT-5.2. We observe 100\% agreement between automated extraction and human judgment, indicating that GPT-5.2 serves as a stable proxy for answer reading in comic-based reasoning. The detail of human verification is in Appendix~\ref{exact_reader_human}.

\subsection{Path II: Comics-as-Context Reasoning.}
In Path~II, comics are used solely as intermediate contextual representations, while the final answer is explicitly generated by a MLLMs in textual form.
Answer extraction in this setting is performed by directly parsing the final model output.
The predicted answer is matched against the ground-truth label using standard normalization and exact-match rules to compute ACC.

% \paragraph{Evaluation Consistency.}
% Both paths are evaluated under identical dataset splits and accuracy metrics. The path-specific answer extraction procedures are designed to reflect the different roles of comics in reasoning, while maintaining consistent evaluation criteria across benchmarks.

\section{Prompt Examples}
\subsection{Prompt in the Main Experiment}

\begin{center}
	%\scriptsize
	\fontsize{8.4}{8.4} \selectfont
	\begin{tcolorbox}[width=1\textwidth, colback=color_prompt, title={\textbf{Prompt for Gemini-3 Pro Image in MATH500 \& GSM8K}}]
        Please help me draw a comic to solve this math problem: \textbf{\{question\}}
	\end{tcolorbox}
\end{center}

\begin{center}
	%\scriptsize
	\fontsize{8.4}{8.4} \selectfont
	\begin{tcolorbox}[width=1\textwidth, colback=color_prompt, title={\textbf{Prompt for Gemini-3 Pro Image in MathVista}}]
        Please draw a suitable comic strip to help solve this math/visual reasoning problem based on the provided image.\\
        
        Context: This is a visual question answering problem at not applicable level, involving natural image.\\
        
        Required skills: numeric commonsense, arithmetic reasoning.\\
        
        Question: \textbf{\{question\}} \\
        
        The answer should be a integer value.\\
        
        Goal: Create a comic that illustrates the problem-solving process step-by-step, showing the complete reasoning from understanding the question to finding the answer.
	\end{tcolorbox}
\end{center}

\begin{center}
	%\scriptsize
	\fontsize{8.4}{8.4} \selectfont
	\begin{tcolorbox}[width=1\textwidth, colback=color_prompt, title={\textbf{Prompt for Gemini-3 Pro Image in CulturalBench-Easy}}]
        Please draw a comic strip to help solve this multiple-choice cultural knowledge question about \textbf{\{country\}}.\\
        
        Question: \textbf{\{question\}}\\
        
        Options:\textbf{\{options\}} \\
        
        Your comic should:\\
        1. Visually depict the cultural scenario described in the question.\\
        2. Show the correct cultural practice/behavior/knowledge of \textbf{\{country\}}.\\
        3. Through the comic story, clearly demonstrate which option (A, B, C, or D) is the correct answer.\\
        4. The comic should lead the viewer to understand and identify the correct choice.\\
        
        Goal: Help the viewer select the correct answer from the four options by illustrating the authentic cultural context of \textbf{\{country\}}.
	\end{tcolorbox}
\end{center}

\begin{center}
	%\scriptsize
	\fontsize{8.4}{8.4} \selectfont
	\begin{tcolorbox}[width=1\textwidth, colback=color_prompt, title={\textbf{Prompt for Gemini-3 Pro Image in CulturalBench-Hard}}]
        Please draw a comic strip to help determine if the following cultural statement about \textbf{\{country\}} is TRUE or FALSE.\\
        
        Question: \textbf{\{question\}}\\
        
        Statement to evaluate: \textbf{\{statement\_to\_judge\}} \\
        
        Your comic should:\\
        1. Visually depict the authentic cultural practice/behavior in \textbf{\{country\}}.\\
        2. Show whether this statement accurately represents the real cultural norm.\\
        3. Through the comic story, clearly demonstrate if this statement is TRUE or FALSE.\\
        4. The comic should help the viewer judge the truthfulness of this cultural claim.\\
        
        Goal: Help the viewer determine TRUE or FALSE by illustrating the actual cultural reality of \textbf{\{country\}}.
	\end{tcolorbox}
\end{center}

\begin{center}
	%\scriptsize
	\fontsize{8.4}{8.4} \selectfont
	\begin{tcolorbox}[width=1\textwidth, colback=color_prompt, title={\textbf{Prompt for Gemini-3 Pro Image in DocVQA}}]
        Please draw a suitable comic strip to help answer this document question based on the provided document image.\\
        
        Task Type: Document Visual Question Answering \textbf{\{question\_types\}}\\
        Question: \textbf{\{question\}}\\
        
        Goal: Create a comic that:\\
        1. Shows the key information extraction process from the document.\\
        2. Highlights the relevant parts of the document that contain the answer.\\
        3. Illustrates the reasoning steps to find the correct answer.\\
        4. Makes the final answer clear through visual storytelling.\\
        
        The comic should help explain how to locate and extract the answer from the document.
	\end{tcolorbox}
\end{center}

\begin{center}
    \fontsize{8.4}{8.4} \selectfont
    \begin{tcolorbox}[
        width=1\textwidth, 
        colback=color_prompt, 
        % boxrule=0.5pt, 
        title={\textbf{Prompt for TWI method: G-Image}}
    ]
        \textbf{Instruction:} You are an expert in writing prompts for text-to-image generation. Based on the following image and textual query, write a precise and detailed prompt to generate a image highly relevant to the query. This image will serve as an auxiliary tool to help resolve the task accurately. Consider composition, style, and detail to ensure practicality. \\

        \textbf{Reasoning Protocol:} Based on the question and the additional synthesized image, let's think step by step, but avoid adding visual descriptions during the reasoning process! \\

        \textbf{Output Format:} End your thinking process with the most appropriate answer in the format "ANSWER: (x)" followed by the choice. \\

        \vspace{0.6em}
        \hrule
        \vspace{0.6em}

        \#\#\# \textbf{Question:} Q \\

        \#\#\# \textbf{Choices:} C \\
        
        \#\#\# \textbf{Prompt Generated:} \\
        
        \textless Extra Image Input\textgreater \\
        
        \textbf{Your Response:}
    \end{tcolorbox}
    \label{G-IMG}
\end{center}

% \begin{center}
% 	%\scriptsize
% 	\fontsize{8.4}{8.4} \selectfont
% 	\begin{tcolorbox}[width=1\textwidth, colback=color_prompt, title={\textbf{Prompt for Gemini-3 Pro Image in eBDtheque (comic translation task)}}]
%         Translate all text in this comic page to Chinese.\\
%         Include: speech bubbles, narration boxes, sound effects, and titles.\\
%         Keep the original tone and style. Preserve humor and cultural context where possible.
% 	\end{tcolorbox}
% \end{center}

\subsection{Prompt in the Analysis Experiment}

\begin{center}
	%\scriptsize
	\fontsize{8.4}{8.4} \selectfont
	\begin{tcolorbox}[width=1\textwidth, colback=color_prompt, title={\textbf{Prompt for Gemini-3 Pro Image in Role-playing Narrative Alignment Experiment}}]
        Please help me draw a \textbf{Slice-of-Life style} comic to solve this math problem: \textbf{\{question\}}\\
        
        Please help me draw a Documentary \textbf{realistic style} picture to solve this math problem: \textbf{\{question\}}
	\end{tcolorbox}
    \label{appendix_style_compare_prompt}
\end{center}

\begin{center}
	%\scriptsize
	\fontsize{8.4}{8.4} \selectfont
    \begin{tcolorbox}[width=1\textwidth, colback=color_prompt, title={\textbf{Prompt for Gemini-3 Pro Image in Ablation on Textual Anchoring Experiment: CulturalBench-Easy}}]
        Please draw a comic strip to help solve this multiple-choice cultural knowledge question about \textbf{\{country\}}.\\
        
        Question: \textbf{\{question\}}\\
        
        Options:\textbf{\{options\}} \\
        
        Your comic should:\\
        1. Visually depict the cultural scenario described in the question.\\
        2. Show the correct cultural practice/behavior/knowledge of \textbf{\{country\}}.\\
        3. Through the comic story, clearly demonstrate which option (A, B, C, or D) is the correct answer.\\
        4. The comic should lead the viewer to understand and identify the correct choice.\\
        \textbf{5. Relies solely on visual storytelling; DO NOT contain any text, speech bubbles, narration boxes, or onomatopoeia.}\\
        
        Goal: Help the viewer select the correct answer from the four options by illustrating the authentic cultural context of \textbf{\{country\}}.
	\end{tcolorbox}
    \label{appendix_text_anchor_prompt}
\end{center}

\begin{center}
	%\scriptsize
	\fontsize{8.4}{8.4} \selectfont
    \begin{tcolorbox}[width=1\textwidth, colback=color_prompt, title={\textbf{Prompt for Gemini-3 Pro Image in Ablation on Textual Anchoring Experiment: CulturalBench-Hard}}]
        Please draw a comic strip to help determine if the following cultural statement about \textbf{\{country\}} is TRUE or FALSE.\\
        
        Question: \textbf{\{question\}}\\
        
        Statement to evaluate: \textbf{\{statement\_to\_judge\}} \\
        
        Your comic should:\\
        1. Visually depict the authentic cultural practice/behavior in \textbf{\{country\}}.\\
        2. Show whether this statement accurately represents the real cultural norm.\\
        3. Through the comic story, clearly demonstrate if this statement is TRUE or FALSE.\\
        4. The comic should help the viewer judge the truthfulness of this cultural claim.\\
        \textbf{5. Relies solely on visual storytelling; DO NOT contain any text, speech bubbles, narration boxes, or onomatopoeia.}\\
        
        Goal: Help the viewer determine TRUE or FALSE by illustrating the actual cultural reality of \textbf{\{country\}}.
	\end{tcolorbox}
\end{center}

% \begin{center}
% 	%\scriptsize
% 	\fontsize{8.4}{8.4} \selectfont
%     \begin{tcolorbox}[width=1\textwidth, colback=color_prompt, title={\textbf{Prompt for Gemini-3 Pro Image in Ablation on Textual Anchoring Experiment: DocVQA}}]
        
% 	\end{tcolorbox}
% \end{center}

\begin{center}
	%\scriptsize
	\fontsize{8.4}{8.4} \selectfont
    \begin{tcolorbox}[width=1\textwidth, colback=color_prompt, title={\textbf{Prompt for Gemini-3 Pro in Structural Coherence Experiment: Global Visual Reasoning}}]
        Please create a complete \{num\_panels\}-panel comic strip that illustrates the step-by-step solution process for this math problem.\\

        Problem: \textbf{\{question\}}\\

        Requirements:\\
        1. Create exactly \{num\_panels\} panels arranged in a coherent sequence.\\
        2. Panel 1: Introduce the problem and key information.\\
        3. Panel 2-\{num\_panels-1\}: Show the logical reasoning steps progressively.\\
        4. Panel \{num\_panels\}: Present the final solution and answer.\\
        5. Maintain consistent characters/elements across all panels.\\
        6. Include clear mathematical notation and explanations in each panel.\\
        7. Ensure smooth visual transitions between panels.\\
        8. Each panel should build logically on the previous one.\\

        Important: Generate ALL \{num\_panels\} panels as a single cohesive comic image with clear panel divisions.
	\end{tcolorbox}
\end{center}

\begin{center}
    \fontsize{8.4}{8.4} \selectfont
    \begin{tcolorbox}[width=1\textwidth, colback=color_prompt,
    title={\textbf{Prompt for Gemini-3 Pro in Structural Coherence Experiment: Incremental Visual Reasoning}}]
    \textbf{Case 1: First Panel ($panel\_num = 1$)}\\
    
    Create a realistic photo-style image (Step 1 of \{total\_panels\}) for solving this math problem.\\
    
    Problem: \textbf{\{Question\}}\\
    
    Style: Realistic photo, NOT cartoon or comic style.\\
    
    This is the FIRST step image. It should:\\
    1. Introduce the problem scenario with realistic visual elements.\\  
    2. Set up real-world objects or scenes that represent the mathematical concepts.\\  
    3. Clearly present the mathematical question in a realistic context.\\  
    4. Use photorealistic rendering, natural lighting, and realistic textures.\\  
    
    Generate ONLY Step 1 as a single realistic photo-style image.
    
    \vspace{0.6em}
    \hrule
    \vspace{0.6em}
    
    \textbf{Case 2: Intermediate Panels ($1 < panel\_num < total\_panels$)}\\
    
    Create a realistic photo-style image (Step \{panel\_num\} of \{total\_panels\}) for solving this math problem.\\
    
    Problem: \textbf{\{Question\}}\\
    
    Style: Realistic photo, NOT cartoon or comic style.\\
    
    This is Step \{panel\_num\} of \{total\_panels\}. It should:\\
    1. Continue logically from the previous step.\\  
    2. Show the next step in the reasoning or calculation process with realistic visuals.\\  
    3. Maintain consistent visual elements from previous image.\\  
    4. Prepare for the next step in the solution.\\  
    5. Use photorealistic rendering, natural lighting, and realistic textures.\\  
    
    Based on the previous image shown, continue the problem-solving process.\\  
    Generate ONLY Step \{panel\_num\} as a single realistic photo-style image.
    
    \vspace{0.6em}
    \hrule
    \vspace{0.6em}
    
    \textbf{Case 3: Final Panel ($panel\_num = total\_panels$)}\\
    
    Create a realistic photo-style image (Step \{panel\_num\} of \{total\_panels\}, the FINAL step) for solving this math problem.\\
    
    Problem: \textbf{\{Question\}}\\
    
    Style: Realistic photo, NOT cartoon or comic style.\\
    
    This is the LAST step (Step {panel\_num} of {total\_panels}). It should:\\
    1. Continue from the previous step's reasoning.\\  
    2. Show the final calculation or conclusion with realistic visual elements.\\  
    3. Present the final answer clearly in a realistic context.\\  
    4. Provide a satisfying conclusion to the problem-solving journey.\\  
    
    Based on the previous image shown, maintain visual consistency and complete the solution.\\  
    Generate ONLY Step \{panel\_num\} as a single realistic photo-style image.
    
    \end{tcolorbox}
\end{center}

\begin{center}
	\fontsize{8.4}{8.4} \selectfont
	\begin{tcolorbox}[width=1\textwidth, colback=color_prompt, title={\textbf{Prompt for GPT-5.2 Answer Extraction in Path I}}]
        You are an answer extraction model.\\
        
        Context: You are given a multi-panel comic that visually depicts a complete reasoning process, together with the original question.\\
        
        Task: Read the comic and identify the final answer shown in the last panel.\\
        
        Rules:
        \begin{itemize}
            \item Only output the final answer.
            \item Do not explain the reasoning.
            \item If the answer is numeric, output the normalized numeric form.
            \item If the answer is a short phrase or option, output it verbatim.
        \end{itemize}
        
        Question: \textbf{\{q\}}
	\end{tcolorbox}
    \label{exact_answer_mllm}
\end{center}

\section{Human Evaluation Protocol}

\subsection{Evaluation for Global and Incremental Visual}
\label{human_evaluation}

We employed three expert annotators to conduct human evaluations for all experiments described in Section~\ref{appendix_reasoing}. All annotators hold a master's degree or higher and have prior experience with vision–language evaluation tasks.

Before annotation, we provided a detailed training session covering task definitions, scoring rubrics, and representative examples. The annotators then completed a pre-annotation phase, during which we aligned interpretations of the evaluation criteria (Accuracy, Logic, State, and Quality) and resolved ambiguities in the scoring guidelines.

Each sample was independently rated by all three annotators. We used the averaged score across annotators as the final reported value. Inter-annotator agreement was monitored throughout the process, and inconsistencies were discussed and resolved according to the established rubric.

\subsection{Evaluation for external answer reader}
\label{exact_reader_human}
To verify the reliability of model-based answer extraction in Path~I, we conduct a manual cross-validation study involving three independent human annotators. A shared subset comprising 20\% of the evaluation instances is randomly sampled across benchmarks. For each sampled instance, all three annotators are provided with the original question and the generated multi-panel comic, and independently identify the final answer depicted in the a panel. The annotations are then compared across annotators to ensure consistency, and any discrepancies are resolved through discussion to reach a consensus. The consensus human answer is finally compared against the answer extracted by GPT-5.2 under identical normalization rules. We observe complete agreement between the consensus human judgments and the automated extraction, supporting the reliability of GPT-5.2 as an answer reader in comic-based reasoning.

\section{Examples of TwC}
This section provides qualitative examples of Thinking with Comics (TwC). It begins with a comparison of different comic styles, followed by illustrative examples from Reasoning Tasks and (Long) Context Understanding Tasks. In total, five benchmarks are included to demonstrate the use of TwC under different task formulations and contextual requirements.

\subsection{Comparison of Different Comic Styles}
\label{appendix_style_compare_example}
We provide qualitative examples of different comic-style visualizations for problem solving. The Documentary style mainly relies on realistic images to directly present the problem context and relevant information. The Role-playing style introduces explicit characters or professional roles, through which the reasoning process is narrated and unfolded in a role-driven manner. In contrast, the Slice-of-life style embeds the reasoning process within everyday scenarios, illustrating problem solving through familiar daily-life activities.

\begin{center}
  \vspace{0.2in}
  \includegraphics[width=\textwidth]{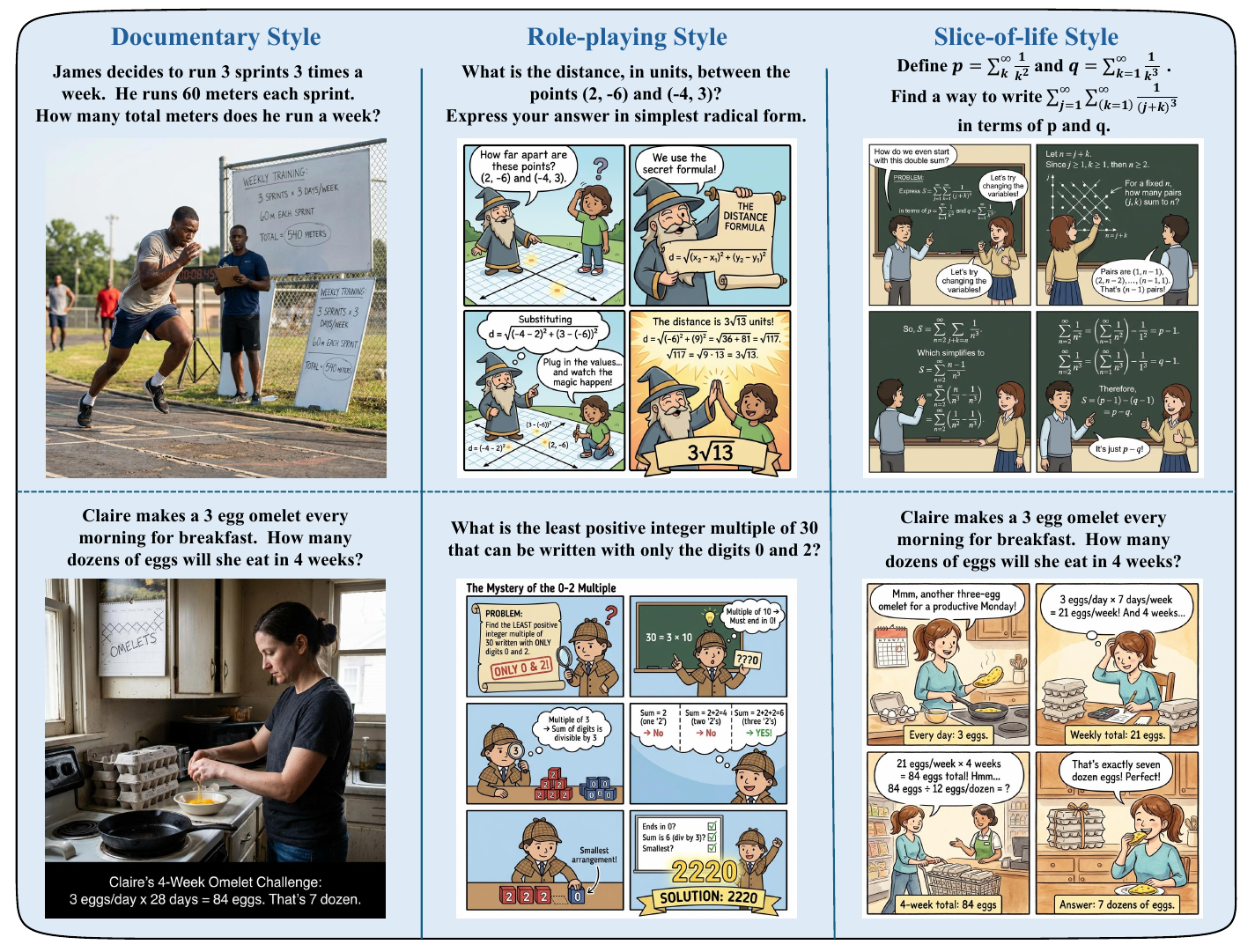}
  \vspace{0.1in}
  \textbf{Figure F1-1:} Examples of different comic-style visualizations for problem solving.
\end{center}

% \begin{figure}[b]
%   \centering
%   \vskip 0.2in
%   \includegraphics[width=\textwidth]{graph/appendix_style_compare0}
%   \caption{
%     Examples of different comic-style visualizations for problem solving.
%   }
%   \label{}
% \end{figure}

\subsection{Reasoning Tasks}

\begin{center}
  \vspace{0.2in}
  \includegraphics[width=\textwidth]{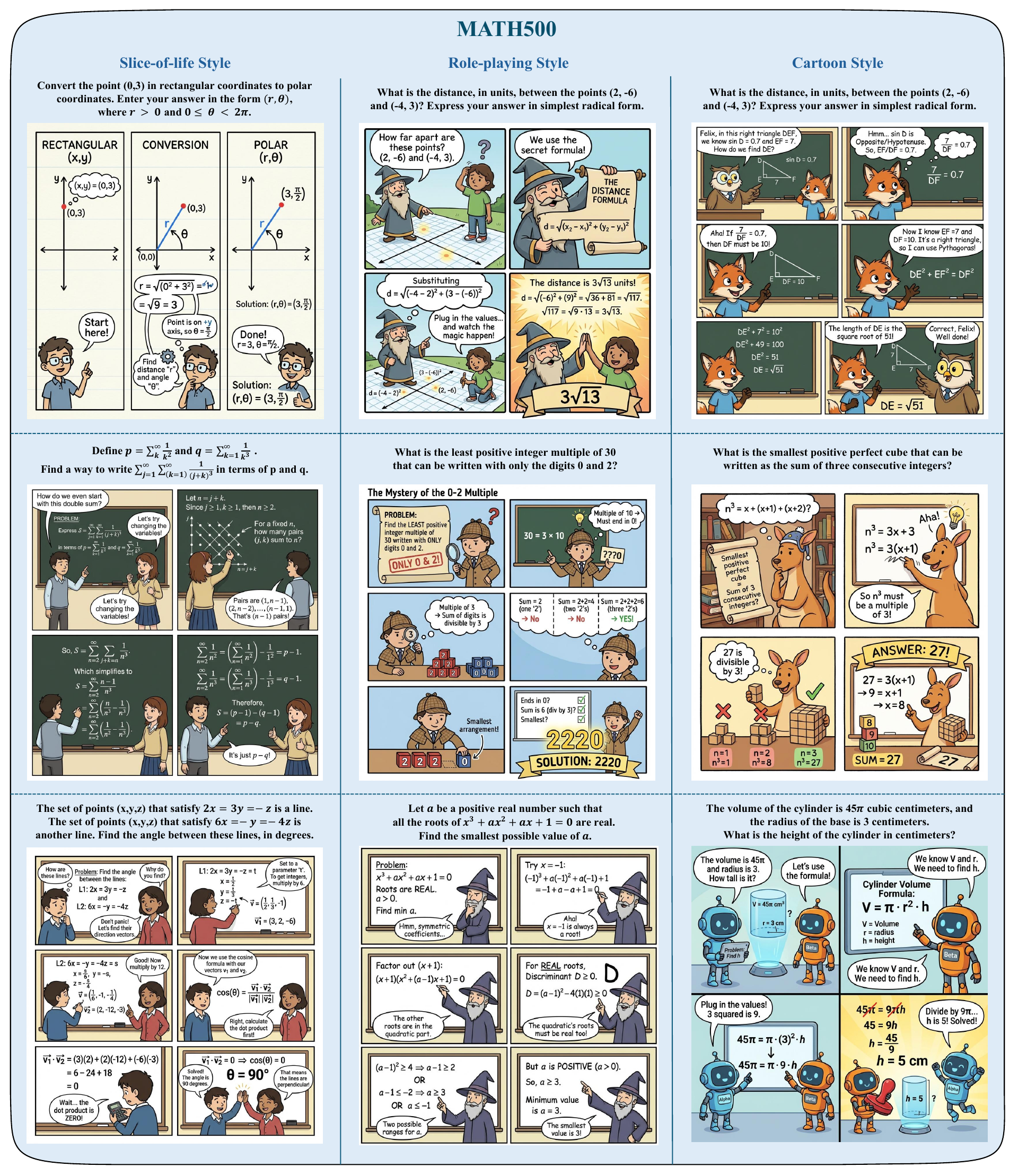}
  \vspace{0.1in}
  \textbf{Figure F2-1:} MATH500
\end{center}

\begin{center}
  \vspace{0.2in}
  \includegraphics[width=\textwidth]{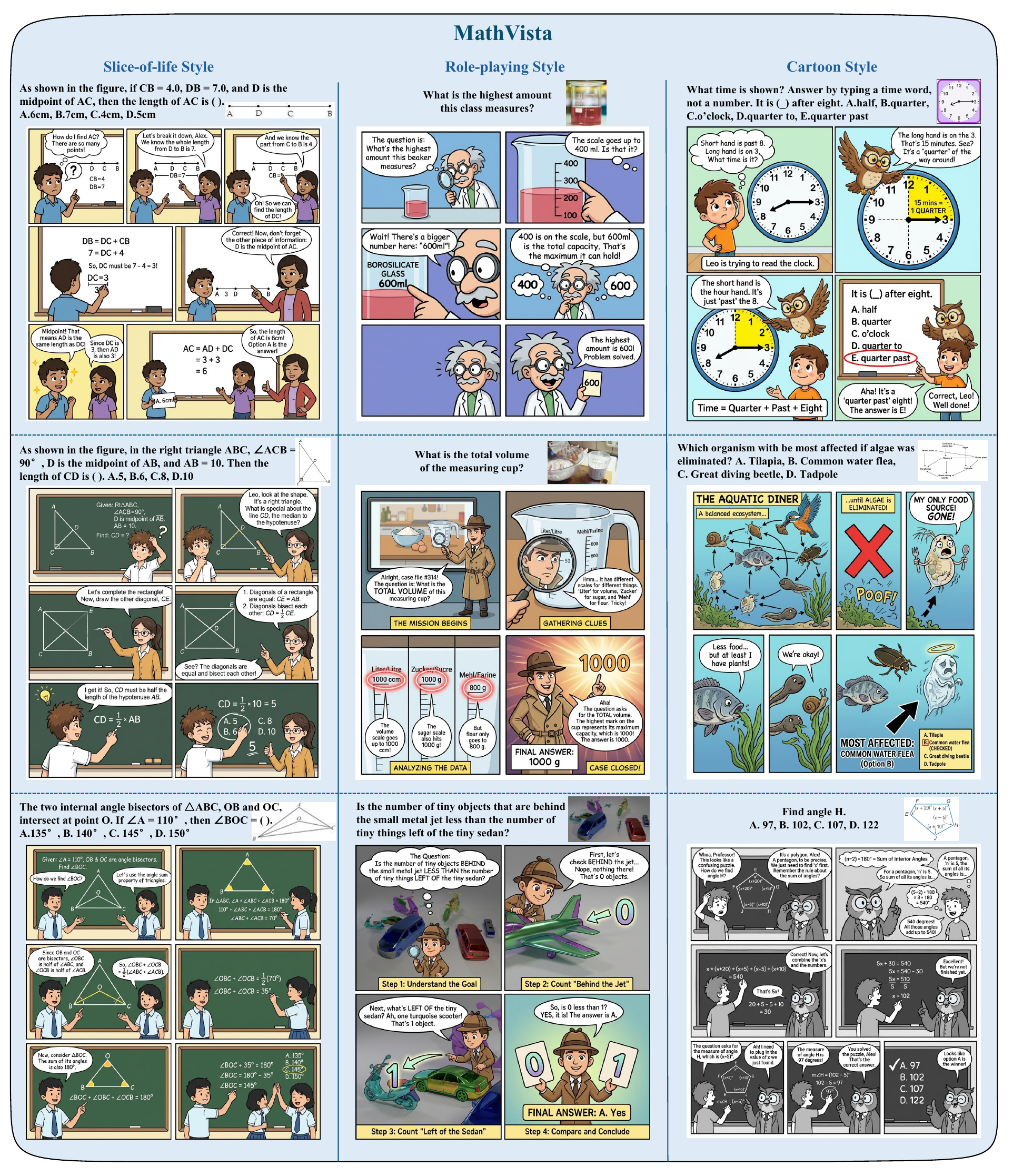}
  \vspace{0.1in}
  \textbf{Figure F2-2:} MathVista
\end{center}

\begin{center}
  \vspace{0.2in}
  \includegraphics[width=\textwidth]{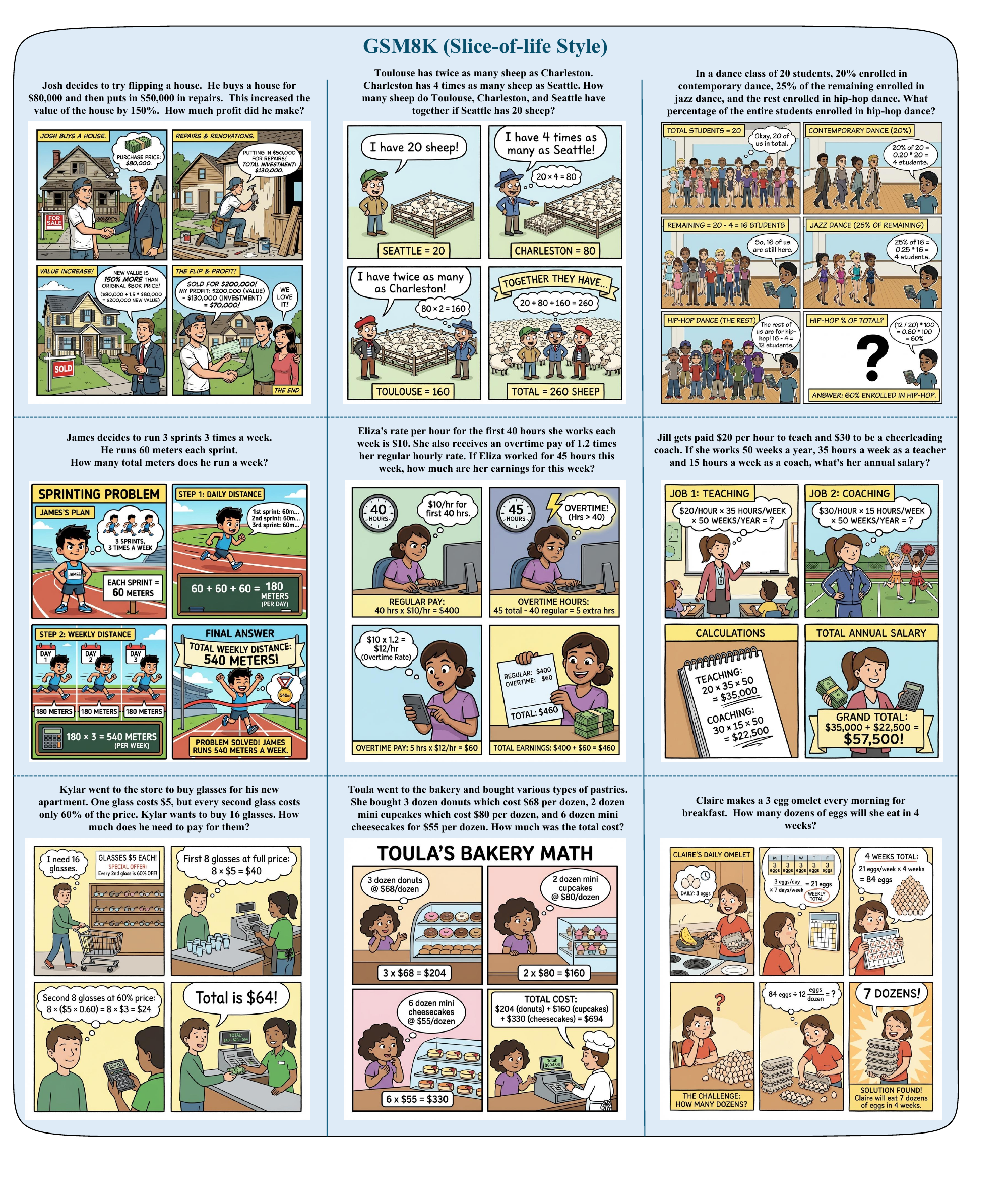}
  \vspace{0.1in}
  \textbf{Figure F2-3:} GSM8K
\end{center}

\subsection{(Long) Context Understanding Tasks}

\begin{center}
  \vspace{0.2in}
  \includegraphics[width=\textwidth]{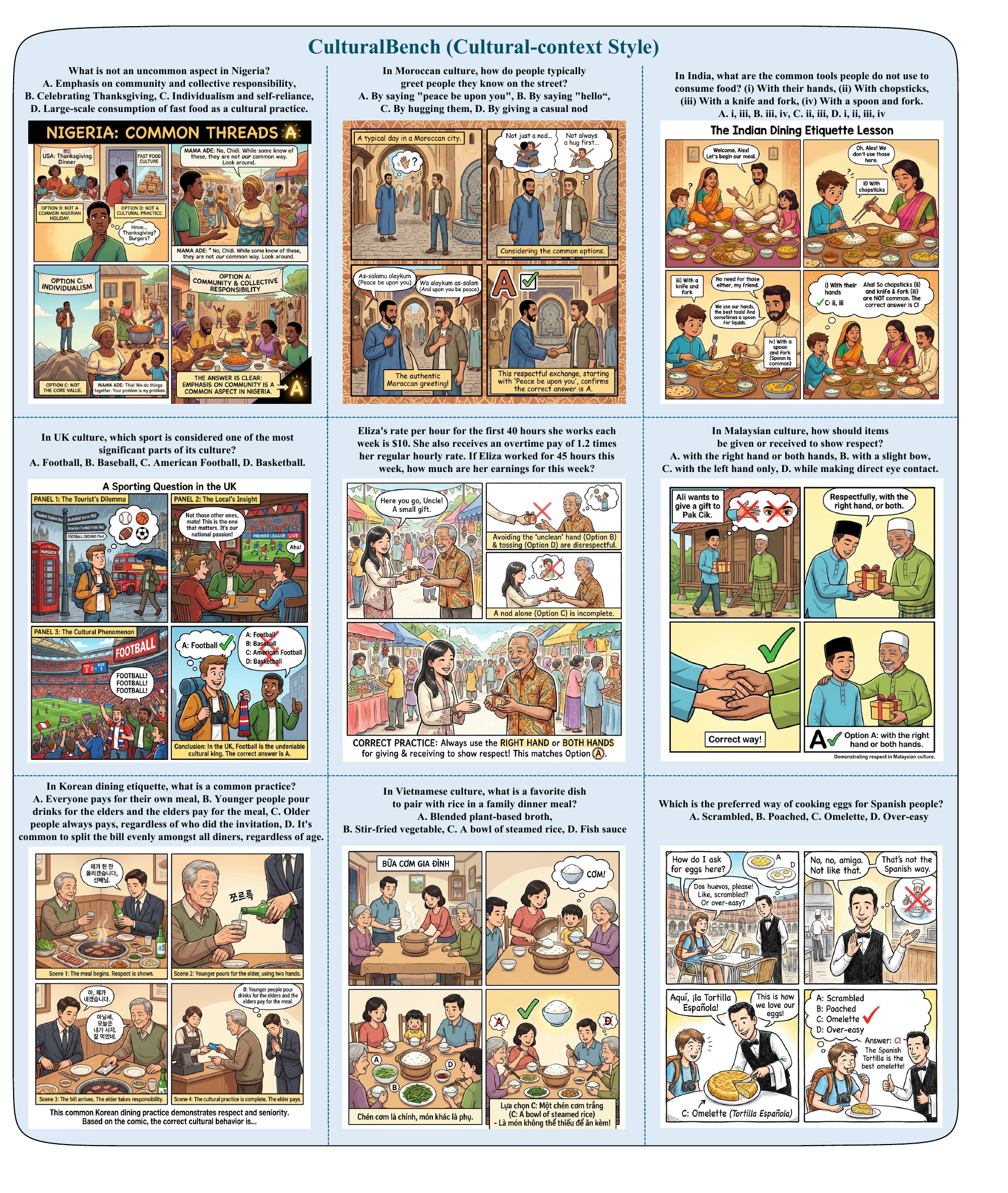}
  \vspace{0.1in}
  \textbf{Figure F3-1:} CulturalBench
\end{center}

\begin{center}
  \vspace{0.2in}
  \includegraphics[width=\textwidth]{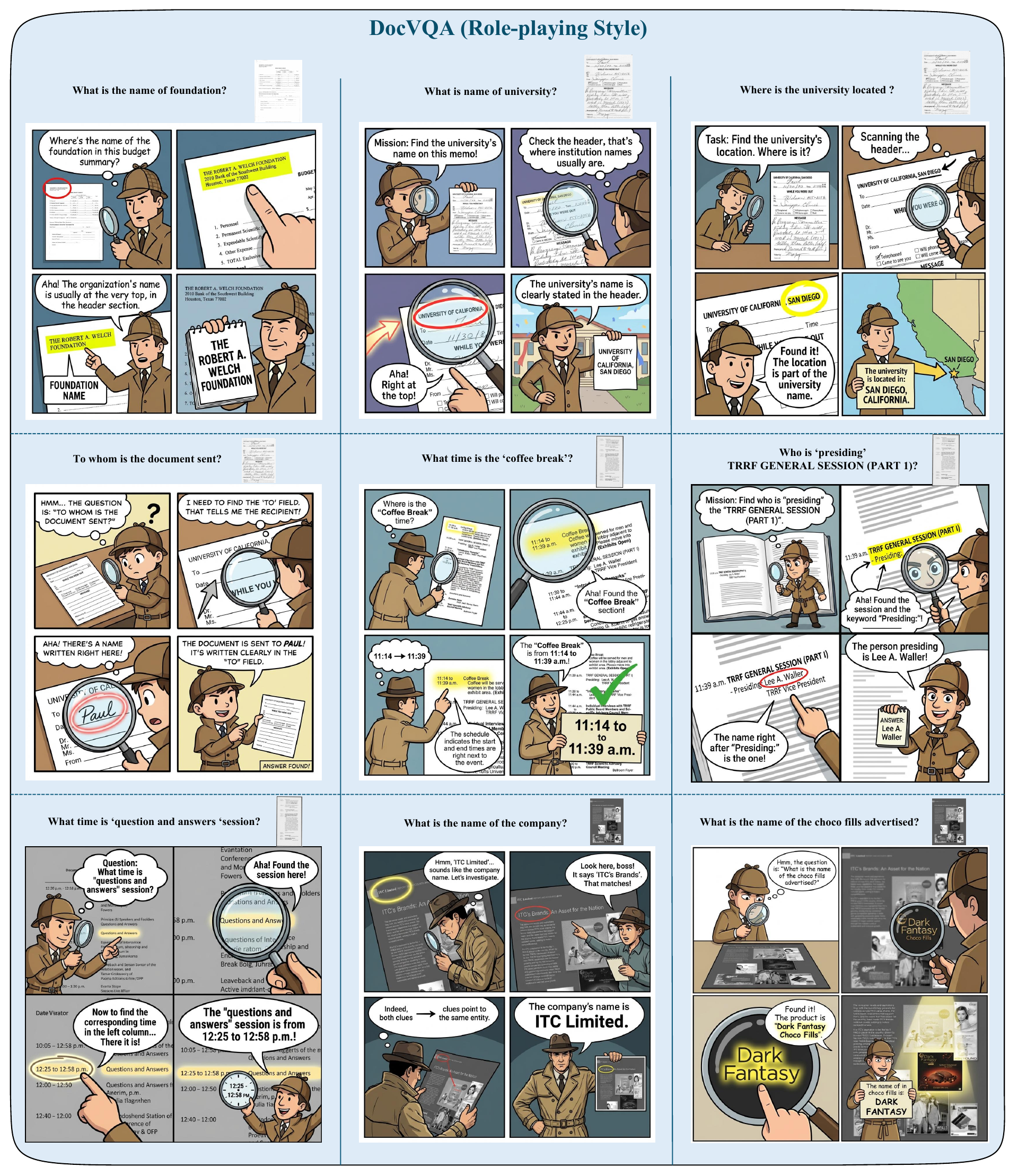}
  \vspace{0.1in}
  \textbf{Figure F3-2:} DocVQA
\end{center}

% \begin{center}
%   \vspace{0.2in}
%   \includegraphics[width=\textwidth]{graph/appendix_eBDtheque}
%   \vspace{0.1in}
%   \textbf{Figure C3-3:} eBDtheque
% \end{center}

% \section{You \emph{can} have an appendix here.}

% You can have as much text here as you want. The main body must be at most $8$
% pages long. For the final version, one more page can be added. If you want, you
% can use an appendix like this one.

% The $\mathtt{\backslash onecolumn}$ command above can be kept in place if you
% prefer a one-column appendix, or can be removed if you prefer a two-column
% appendix.  Apart from this possible change, the style (font size, spacing,
% margins, page numbering, etc.) should be kept the same as the main body.
%%%%%%%%%%%%%%%%%%%%%%%%%%%%%%%%%%%%%%%%%%%%%%%%%%%%%%%%%%%%%%%%%%%%%%%%%%%%%%%
%%%%%%%%%%%%%%%%%%%%%%%%%%%%%%%%%%%%%%%%%%%%%%%%%%%%%%%%%%%%%%%%%%%%%%%%%%%%%%%

\end{document}